\documentclass{article}

\usepackage{microtype}
\usepackage{graphicx}
\usepackage{subfigure}
\usepackage{booktabs} 

\usepackage{hyperref}

\usepackage{mathtools}
\usepackage{xcolor}
\usepackage[frozencache,cachedir=.]{minted}
\usepackage{multirow}
\usepackage{multicol}
\usepackage{nicefrac}

\newcommand{\round}[1]{\ensuremath{\lfloor#1\rceil}}

\usepackage[accepted]{icml2021}

\icmltitlerunning{Variable-rate discrete representation learning}

\begin{document}

\twocolumn[
\icmltitle{Variable-rate discrete representation learning}



\icmlsetsymbol{equal}{*}

\begin{icmlauthorlist}
\icmlauthor{Sander Dieleman}{dm}
\icmlauthor{Charlie Nash}{dm}
\icmlauthor{Jesse Engel}{gb}
\icmlauthor{Karen Simonyan}{dm}
\end{icmlauthorlist}

\icmlaffiliation{dm}{DeepMind, London, UK}
\icmlaffiliation{gb}{Google Research, Brain Team, Mountain View, CA, USA}

\icmlcorrespondingauthor{Sander Dieleman}{sedielem@google.com}

\icmlkeywords{Deep learning, discrete representations, representation learning, variable-rate, generative modelling, language modelling, speech, audio}

\vskip 0.3in
]



\printAffiliationsAndNotice{}  

\begin{abstract}
Semantically meaningful information content in perceptual signals is usually unevenly distributed. In speech signals for example, there are often many silences, and the speed of pronunciation can vary considerably. In this work, we propose \emph{slow autoencoders} (SlowAEs) for unsupervised learning of high-level variable-rate discrete representations of sequences, and apply them to speech. We show that the resulting \emph{event-based} representations automatically grow or shrink depending on the density of salient information in the input signals, while still allowing for faithful signal reconstruction. We develop \emph{run-length Transformers} (RLTs) for event-based representation modelling and use them to construct language models in the speech domain, which are able to generate grammatical and semantically coherent utterances and continuations. Samples can be found at \url{https://vdrl.github.io/}.
\end{abstract}

\section{Introduction}
\label{sec:introduction}

\begin{figure}[t]
\vskip 0.2in
\begin{center}
\centerline{\includegraphics[width=\columnwidth]{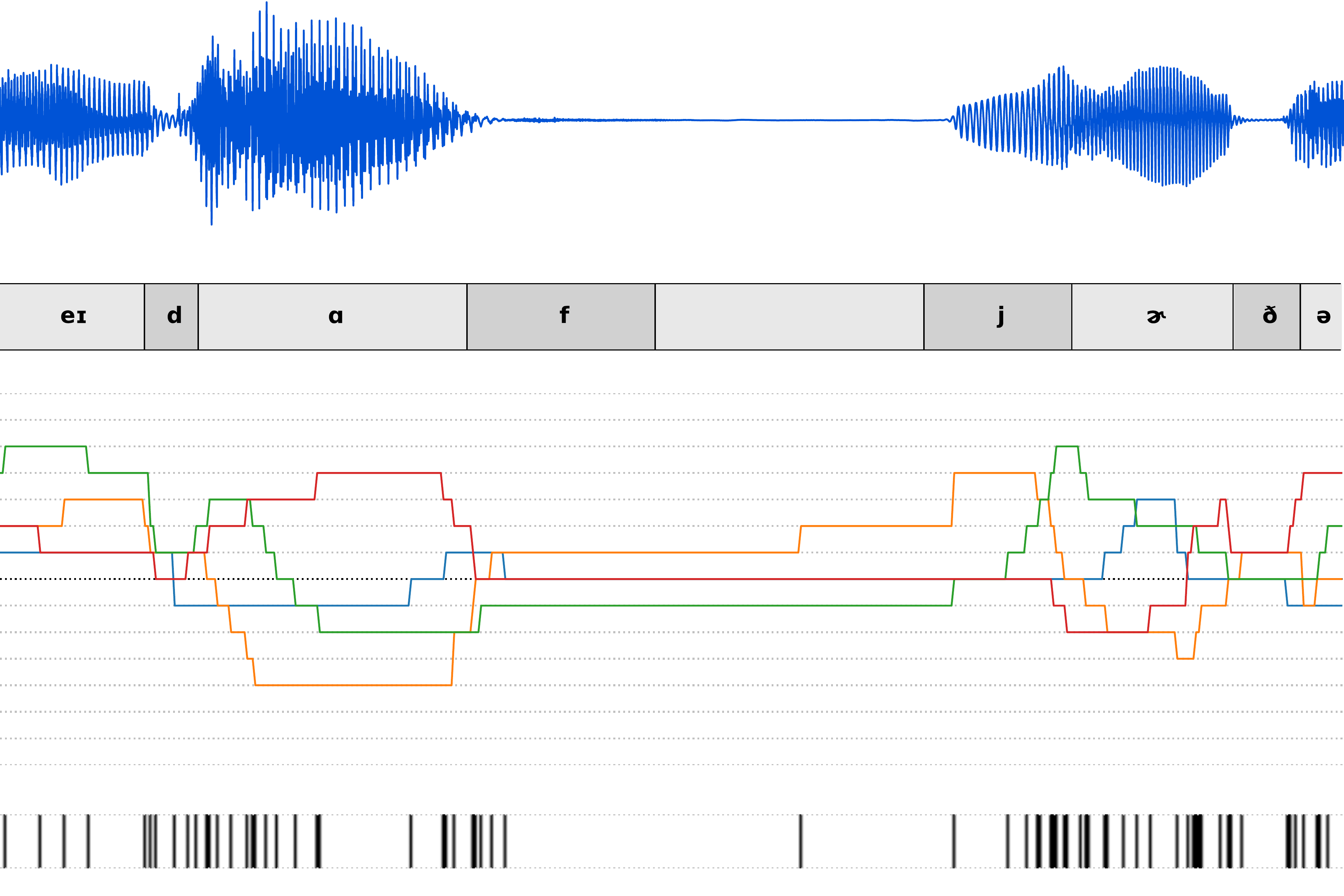}}
\caption{From top to bottom: (a) a waveform corresponding to one second of American English speech: \emph{`[... has p]aid off! You're the [...]'}; (b) the corresponding aligned phoneme sequence (using the International Phonetic Alphabet); (c) a discrete representation of the speech segment with 4 channels, learnt with the proposed \emph{slow autoencoder} (SlowAE) model from audio only (without supervision), which is amenable to run-length encoding; (d) a `barcode' plot indicating where events (changes in the discrete representation) occur. Event density is correlated with meaningful signal structure, e.g. changes in phonemes.}
\label{fig:firstpage}
\end{center}
\vskip -0.2in
\end{figure}

The density of semantically meaningful information in perceptual signals (e.g. audio, images and video) usually varies considerably over time and space. For example, in speech signals, there are often many silences, and some utterances may be pronounced more slowly and deliberately than others. In images, large groups of pixels may consist solely of background textures, while other, much smaller groups of pixels encode salient features such as edges and corners.

Digital representations of these signals are typically \emph{fixed-rate}: they use a fixed number of bits per unit of data (e.g. per audio sample, image pixel, video frame). This means that the same amount of information is always encoded, regardless of whether it is perceptually salient: two seconds of speech and two seconds of silence are represented with the same number of bits. Compression algorithms represent signals more efficiently by removing information that is not perceptually salient (lossy compression) and capturing redundancy in the signals. When applied to signals with an uneven distribution of salient content, this results in \emph{variable-rate} representations that allocate more bits to regions with content.

Compression algorithms are useful for efficient storage and transmission of perceptual signals, but compressed representations tend to be less practical for information retrieval (e.g. speech recognition, image classification) and modelling, especially with machine learning models. This is because compression tends to obfuscate meaningful structure, which we could otherwise exploit to endow our models with powerful inductive biases. This is especially the case when entropy coding is used, and that happens to be a core component of virtually all modern compression algorithms.

It is therefore unsurprising that current popular machine learning methods for representation learning usually operate on uncompressed, structured signal representations (e.g. PCM\footnote{Pulse-code modulation}-encoded audio, pixel grids), and produce fixed-size higher-level representations. This can be wasteful, as it implies that the same amount of computation is always used when these representations are processed for downstream tasks, regardless of their salient information content.

\subsection{Event-based representations}
\label{sec:event-based}

An alternative strategy to represent a perceptual signal more efficiently is to use an \emph{event-based} approach: only encode interesting changes in the nature of the signal, and their position relative to other changes. In other words, we can irregularly sample the signal, rather than at fixed intervals or on a fixed grid. We can then construct a representation whose size is automatically adapted to the amount of perceptually relevant information in the signal.

The size of an event-based representation is adaptive in much the same way as that of a fixed-rate representation after compression. In the context of machine learning, this leads to the prospect of computation and memory requirements that scale with the amount of salient content in a signal. For example, with an event-based representation of speech, a model could use much less computation to process two seconds of silence compared to two seconds of speech, because it requires fewer events to represent. The question is how to obtain such event-based representations from data, and how to process them in a way that matches the performance of dense representations. One possible strategy is to use \emph{run-length encoding}~\cite{rle}, which encodes runs of constant values in an input signal as $(\mathrm{value}, \mathrm{length})$-tuples.

What constitutes an interesting \emph{event} (i.e. a meaningful change in the nature of the signal) is highly domain-dependent and potentially difficult to narrow down. In a speech signal, for example, it could be the point at which the speaker transitions from pronouncing one phoneme to another. However, there may be other meaningful changes that do not necessarily align with phoneme changes (e.g. changes in volume or pitch), so having to manually determine which events to record in the representation may not be practical. Therefore, we endeavour to construct an unsupervised learning algorithm that automatically identifies such events from data.

\subsection{Contributions}

In this work, we develop \textbf{slow autoencoders} (SlowAEs), which learn slowly varying discrete representations of input signals. We use an adaptive group-sparse slowness penalty and a novel scalar quantisation strategy to make the encoder outputs amenable to run-length encoding, yielding event-based representations.

We also design \textbf{run-length Transformers} (RLTs), which autoregressively model these discrete event-based representations efficiently by exploiting their underlying structure.

Finally, we use the combination of slow autoencoders and run-length Transformers to build a \textbf{language model in the speech domain}, trained on event-based representations of a large collection of audiobook recordings. The model produces new utterances that are grammatical and sometimes meaningful, without requiring any form of supervision.

In what follows, we first discuss discrete representation learning (\S\ref{sec:discrete-representation-learning}) and argue our focus on discrete, rather than continuous representations. We then describe our proposed method for learning and modelling discrete event-based representations (\S\ref{sec:variable-rate-discrete-representations}), and examine language models in the speech domain, as well as their relation to more traditional text-based language models (\S\ref{sec:language-modelling-in-the-speech-domain}). Related work (\S\ref{sec:related-work}) is then followed by a description of our experiments and results (\S\ref{sec:experiments-and-results}), and finally a discussion thereof (\S\ref{sec:discussion}).

\section{Discrete representation learning}
\label{sec:discrete-representation-learning}

For many modalities, discrete representations are a very natural fit. For example, language is conceptually a sequence of symbols from a discrete alphabet, and music can in many cases be symbolically represented using musical notation. They are also well-suited for complex reasoning and planning~\cite{alphago,vqvae}.

In the context of machine learning, they have a few significant advantages. For one, discrete representations typically offer a substantially reduced capacity compared to continuous representations. This limit on information content can be a convenience, especially in the context of generative modelling. Applying generative models directly to standard digital representations of perceptual data (e.g. audio waveforms, image pixel grids) usually results in excessive allocation of model capacity towards perceptually irrelevant details~\cite{ham}, whereas models that operate on compressed discrete representations can focus on the statistical dependencies in the data at a more salient level~\cite{dctransformer}. We can also use learning approaches to obtain such representations~\cite{vqvae}. Combining learnt discrete representations with powerful autoregressive priors has been shown in recent years to be a particularly effective recipe for generative modelling~\cite{vqvae2,challenge,jukebox}.

Another advantage of discrete representations is that their capacity is intrinsically limited by the number of available bits. The rate of a representation can be varied in a straightforward manner e.g. by choosing the number of available channels or the number of possible discrete values. For continuous representations, limiting capacity is more involved, and the limit is typically softly enforced during optimization (e.g. the KL penalty in variational autoencoders~\cite{vae1,vae2}).

\subsection{Vector quantisation}
\label{sec:vector-quantisation}

The most popular method for discrete representation learning is the vector quantised variational autoencoder~(VQ-VAE, \citet{vqvae}). This is an autoencoder model with a discrete bottleneck based on vector quantisation (VQ, \citet{quantization}). Encoder outputs are clamped to the corresponding nearest entries in a codebook. This codebook is fit using a K-means-like algorithm, jointly with the other model parameters. The resulting \emph{codes} can be represented as a sequence of integers, and modelled using categorical distributions.

Although the quantisation operation is not differentiable, backpropagation can proceed simply by ignoring it: we propagate the gradients unchanged from the quantized encoder outputs to the original outputs. This approach (\emph{straight-through estimation}, \citet{straightthrough}) yields biased gradient estimates for the encoder, but is effective nevertheless. Alternative strategies for backpropagating through categorical discrete bottlenecks also exist~\cite{gumbelsoftmax,concrete,dvae}. Categorical discrete bottlenecks enable maximal flexibility of the representation: each code can correspond to a particular feature of the input signal, independent of the others.

\subsection{Scalar quantisation}
\label{sec:scalar-quantisation}
An alternative strategy to introduce a discrete bottleneck into a model is through scalar quantisation, e.g. rounding encoder outputs to the nearest integer. This approach is more commonly used in learned image compression models~\cite{compressionballe,compressiontheis}. There is no need to learn a codebook in this setting, which can be a significant advantage as it is a source of instability when training models with VQ bottlenecks.

Rounding to the nearest integer implies an ordinal relationship between different codes, however, which means they cannot easily represent completely disparate features in the input signal. This can be compensated for to some degree by designing bottlenecks with multiple channels, but nevertheless, more bits will be required to represent data with comparable fidelity to VQ. That said, ordinality can also be a useful inductive bias~\cite{pianogenie}.

\subsection{Learning representations}
\label{sec:learning-representations}

We will focus on the autoencoder paradigm for representation learning, because we are interested in generative modelling and wish to map any generated signals back to the input domain. We note however that discrete bottlenecks can also be paired with other network architectures and loss functions~\cite{vqwav2vec,ham,vqcpc}.

The decoder model can have a significant impact on the kind of information that gets encoded in the learnt representations. Feed-forward decoders are commonly used due to their relative simplicity~\cite{vqvae2,jukebox}, but in this work, we use autoregressive decoders instead; they can model the bulk of the local structure present in a signal without making use of the encoder, and thus allow the latter to focus on capturing more high-level structure~\cite{lossy}. This is desirable, as our goal is to learn compact high-level representations that make abstraction of local details.

The learnt representation typically has a much lower capacity than the original input representation. This is generally achieved through gradual resolution reduction in the encoder (e.g. with pooling layers). As a result, every code captures information about a particular region of the input, and all these regions are the same size. Not only does this make it difficult for the model to use the bulk of its representational capacity where it is most needed (i.e. for regions of the input with more perceptually salient content); it also means that some parts of the representation will encode irrelevant information, because there is nothing else for them to capture. This in turn makes models that process these representations less able to use their capacity efficiently. This is a key limitation, which we will try to address next.

\section{Variable-rate discrete representations}
\label{sec:variable-rate-discrete-representations}

To learn more efficient discrete representations, we propose to make them \emph{event-based}, so that more capacity can be used to describe more perceptually salient parts of the input.
We will first develop a discrete autoencoder to extract such representations, and then design a Transformer architecture to model the representations autoregressively.

\subsection{Slow autoencoders (SlowAEs)}
\label{sec:slowaes}

We wish to construct an autoencoder model with a discrete bottleneck, where the encoder produces discrete sequences that are amenable to \emph{run-length encoding} (RLE): $l$ repeated instances of the same value $v$ will be represented as a $(v,l)$-tuple. We will refer to sequences of such tuples as \emph{event sequences}, because each element (\emph{event}) indicates a change in value in the original sequence. If such changes are infrequent enough, the event sequence will be a more compact representation of the same information.

\begin{figure}[t]
\vskip 0.2in
\begin{center}
\centerline{\includegraphics[width=\columnwidth, trim=0 110 420 0, clip]{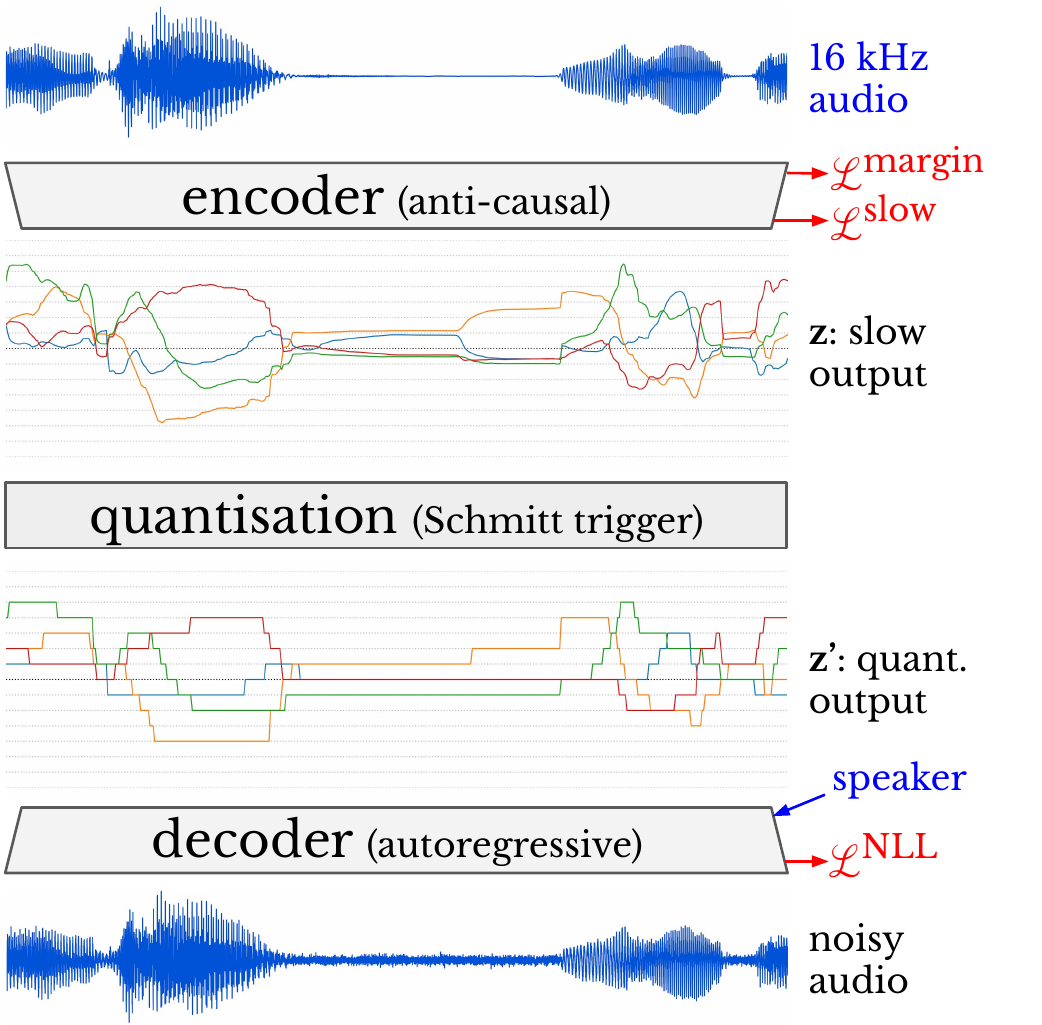}}
\caption{Schematic overview of the SlowAE architecture. Inputs are {\color{blue} blue}, loss terms are {\color{red} red}. The encoder $e_\phi$ maps mu-law encoded audio at $16 \mathrm{kHz}$ to a multi-channel slowly varying representation $\mathbf{z}$ at $500 \mathrm{Hz}$. The quantiser $q$ turns this into a discrete representation $\mathbf{z'}$. The decoder $d_\theta$ autoregressively predicts a noisy version of the input signal, conditioned on $\mathbf{z'}$ and the speaker identity. $\mathcal{L}^\mathrm{margin}$ is applied to $\mathbf{z}$ to encourage it to stay within the range $[-1, 1]$. $\mathcal{L}^\mathrm{slow}$ encourages $\mathbf{z}$ to vary slowly over time. $\mathcal{L}^\mathrm{NLL}$ is the audio reconstruction loss term. A more detailed diagram of the SlowAE architecture is shown in Figure~\ref{fig:slowae-diagram}.}
\label{fig:slowae}
\end{center}
\vskip -0.2in
\end{figure}

Let $\mathbf{x}$ be an input signal, $e_\phi$ an encoder network parameterised by $\phi$, and $d_\theta$ a decoder network parameterised by $\theta$. $q$ is a quantisation function, for which we will obtain approximate gradients using straight-through estimation. We use $\mathbf{z} = e_\phi(\mathbf{x})$ to represent the corresponding encoder output before quantisation, and $\mathbf{z'} = q(\mathbf{z})$ for the quantised output. The model, which we name \emph{slow autoencoder} (SlowAE), will be trained to maximise the reconstruction log-likelihood of the input signal under the distribution induced by the decoder $p_{d_\theta}(\mathbf{x} | \mathbf{z'})$: $\mathcal{L}^\mathrm{NLL} = -\log p_{d_\theta}(\mathbf{x} | \mathbf{z'})$ (negative log-likelihood). $\mathbf{z}$ and $\mathbf{z'}$ are sequences of $T$ vectors of size $C$, where $T$ is the number of sequence steps and $C$ the number of encoder output channels. A diagram of the SlowAE is shown in Figure~\ref{fig:slowae}.

To limit the capacity of the bottleneck explicitly, we use a scalar quantisation function $q$. We choose a number of quantisation levels and impose a margin penalty to discourage the encoder to output values outside the valid range. With $2k + 1$ levels, we scale the encoder output $\mathbf{z}$ by $k$ before quantisation, so that the valid range of the unscaled output is always $[-1, 1]$. Following \citet{pianogenie}, we add a loss term $\mathcal{L}^\mathrm{margin}(\mathbf{z}) = \sum \max \left(|\mathbf{z}| - 1, 0\right)^2 $, which is non-zero only when $\mathbf{z}$ is not in the valid range. This yields the following loss function:

\begin{equation}
    \mathcal{L} = \mathcal{L}^\mathrm{NLL} + \mu \mathcal{L}^\mathrm{margin} ,
\end{equation}

where we normalise $\mathcal{L}^\mathrm{NLL}$ so it is measured in nats per time step. We find that training is fairly robust to the choice of $\mu$ across a wide range of models, as long as it is large enough; we use $\mu = 100$ for all models.

For computational convenience, the encoder $e_\phi(\mathbf{x})$ incorporates a few strided layers so that the output $\mathbf{z}$ has a rate of $500 \mathrm{Hz}$, which is a temporal resolution reduction of $32\times$ compared to the input signals. This resolution is more than sufficient to capture meaningful high-level changes in speech signals, but low enough to make further processing more practical. For run-length encoding, it also reduces the required resolution for the lengths $l$.

Note that we opt for scalar quantisation instead of vector quantisation, which may seem like a more natural choice at first. This is because next, we will be imposing additional constraints to encourage the encoder to extract slowly varying representations. This is much more straightforward in the ordinal setting than in the categorical one\footnote{\citet{chorowski-slowvq} note that imposing such constraints in the categorical setting can lead to collapse of the latent space.}.

\subsubsection{Slowness penalty}
\label{sec:slowness-penalty}
To be able to represent the sequences produced by the encoder with as few events as possible, they should change slowly over time. We can achieve this by imposing a \emph{slowness penalty} $\mathcal{L}^\mathrm{slow}$ on $\mathbf{z}$. The full loss function is then:

\begin{equation}
    \mathcal{L} = \mathcal{L}^\mathrm{NLL} + \mu \mathcal{L}^\mathrm{margin} + \lambda \mathcal{L}^{\mathrm{slow}} ,
\end{equation}
where the hyperparameter $\lambda$ controls the trade-off between reconstruction quality and slowness of the learnt discrete representation $\mathbf{z'}$. Note that we apply this penalty to $\mathbf{z}$ and not $\mathbf{z'}$, to reduce our dependency on approximate gradients from straight-through estimation.

We consider a few different variants, the simplest of which is the L2 slowness penalty:

\begin{equation}
    \mathcal{L}_\mathrm{L2}^{\mathrm{slow}}(\mathbf{z}) = \frac{1}{(T-1)C} \sum_{t=1}^{T-1} \sum_{c=1}^C \left(z_{t+1,c} - z_{t,c}\right)^2 .
\end{equation}
This penalises the square of the difference between subsequent steps. It is closely related to the objective used in \emph{slow feature analysis}~\cite{sfa}. More recently, \citet{deltavae} and \citet{slowvae} used a probabilistic version~\cite{mlsfa} to limit bottleneck capacity in variational autoencoders.

To encourage slowly changing $\mathbf{z'}$, it is preferable for $\mathbf{z}$ to change in bursts, rather than gradually over time in smaller increments. The L1 slowness penalty achieves this by encouraging sparsity:
\begin{equation}
    \mathcal{L}_\mathrm{L1}^{\mathrm{slow}}(\mathbf{z}) = \frac{1}{(T-1)C} \sum_{t=1}^{T-1} \sum_{c=1}^C \left|z_{t+1,c} - z_{t,c}\right| .
\end{equation}
To minimise this penalty, the differences between subsequent steps should be mostly zero, but occasional larger deviations are acceptable. This is similar to the objective used in \emph{total variation denoising}~\cite{tvd}. In practice however, we found that some channels will typically collapse to zero using this penalty, which negatively affects reconstruction quality (see \S\ref{sec:experiments-slowness-penalties}).
We can address this by constructing a penalty that acts as L1 along the sequence dimension, and as L2 along all other dimensions:

\begin{equation}
    \mathcal{L}_\mathrm{GS}^{\mathrm{slow}}(\mathbf{z}) = \frac{1}{(T-1)C} \left( \sum_{t=1}^{T-1} \sqrt{\sum_{c=1}^C \left(z_{t+1,c} - z_{t,c}\right)^2} \right)^2 .
\end{equation}

We call this the \emph{group-sparse} slowness penalty, because it combines the differences between subsequent steps across all channels into groups, and enforces sparsity across those groups, but not within them. As a side effect, this formulation encourages changes in $\mathbf{z'}$ to happen synchronously across channels. Group sparsity has previously been used for representation learning by \citet{topographic}. The effect of different slowness penalties is visualised in Figure~\ref{fig:compare-codes}. We will use the group-sparse penalty throughout this work, unless otherwise indicated.

\subsubsection{Schmitt trigger quantisation (STQ)}
\label{sec:stq}
We can quantise the slowly varying encoder output $\mathbf{z}$ by rounding appropriately scaled values to the nearest integer: $q(\mathbf{z}) = \frac{1}{k} \round{k \mathbf{z}}$. If $\mathbf{z}$ is limited to $[-1, 1]$, this results in $2k + 1$ quantisation levels. However, this approach has a significant downside: due to noise, these values occasionally straddle the boundary between quantisation levels, and this leads to $\mathbf{z'}$ rapidly jumping back and forth between those levels. This means that many more events will be required to describe the resulting sequences after run-length encoding.

We can address this by using a quantisation scheme with memory: at any point in the sequence, we will only change the quantisation level if the input has moved away from the level at the previous step by more than a given margin $m$. In practice, this means that the quantisation level at step $t$ is defined recursively as follows:

\begin{equation}
    q_m(\mathbf{z}_t) = 
     \begin{cases}
     q_m(\mathbf{z}_{t-1})  & \text{if $\left| q_m(\mathbf{z}_{t-1}) - \mathbf{z}_t \right| \leq m$} \\
     \frac{1}{k} \round{k\mathbf{z}_t} & \text{otherwise}
  \end{cases} .
\end{equation}

For $m \leq \frac{1}{2k}$, this reduces to memoryless quantisation. For higher values, quantisation levels become `sticky'. We use $m = \frac{1}{k}$ and refer to this scheme as \emph{Schmitt trigger quantisation} (STQ), because it behaves like the Schmitt trigger circuit~\cite{schmitttrigger}. The increased margin removes the effect of small fluctuations on the output. It also increases the quantisation error on average, but we find that straight-through estimation still works well in practice.

STQ has proven essential to learn useful representations. Figure~\ref{fig:compare-codes} shows how models without it learn representations that are much less efficient.

\subsubsection{Automatic penalty weighting}
\label{sec:automatic-penalty-weighting}

The average length of event sequences produced by $e_\phi$ depends on the value of $\lambda$, which trades off reconstruction quality and slowness. We refer to the ratio of the number of events and the length of the input sequences as the \emph{average event rate} (AER). Ideally, the AER of the learnt representation should be as low as possible while still allowing for faithful reconstruction of the input signals.

Manually tuning $\lambda$ to target a desired rate is challenging because it is quite sensitive and depends on other aspects of the model, such as the decoder architecture, the type of slowness penalty, and even the learning rate. Therefore, we adapt $\lambda$ on the fly during training, to achieve a target AER~\cite{geco}. This also enables a comparison of different model variants on equal footing, because the capacity of the representation is approximately constant.

We can estimate the AER from $\mathbf{z'}$ by counting the number of changes in value. This quantity is not differentiable, but the relationship between $\lambda$ and the AER is approximately monotonically decreasing. If we want the AER to increase, we can decrease $\lambda$ by a small amount, and vice versa.

\begin{figure}[t]
\vskip 0.2in
\begin{center}
\centerline{\includegraphics[width=\columnwidth]{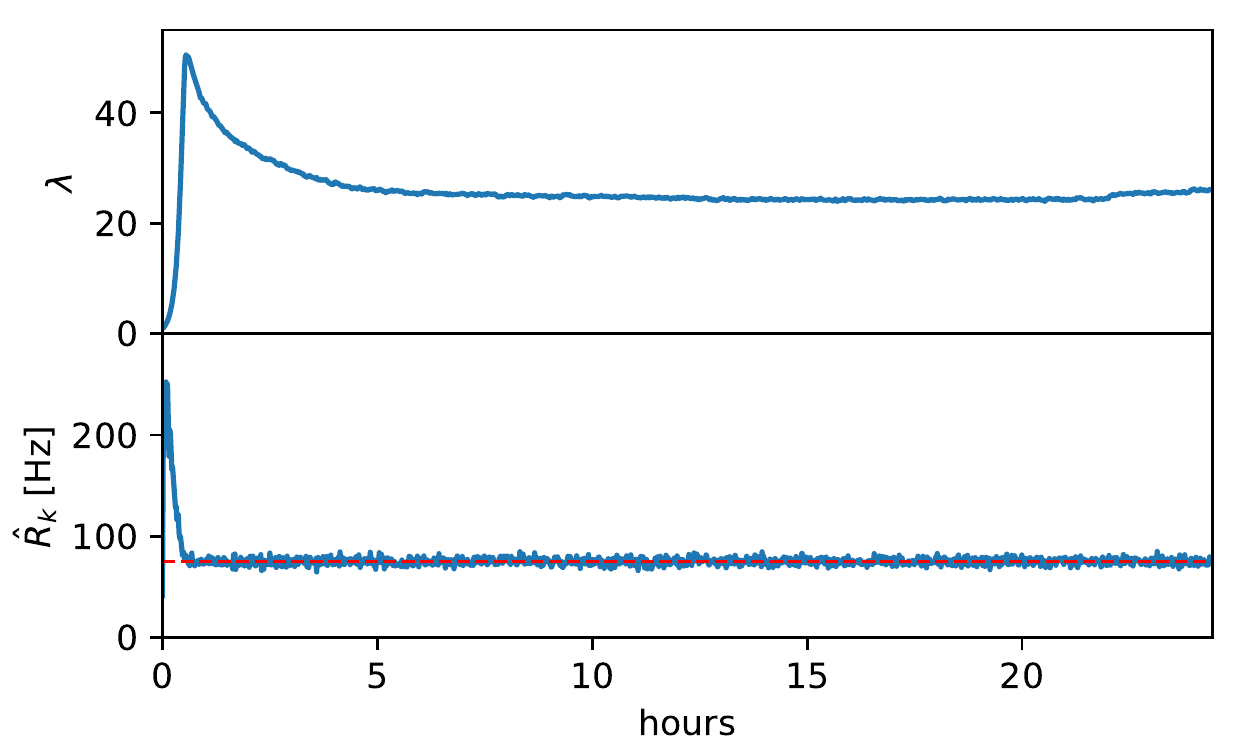}}
\caption{Evolution of $\lambda$ and the average event rate (AER) estimate $\hat{R}_k$ over the course of SlowAE training. The dashed red line indicates the target AER $R_\mathrm{T} = 75\mathrm{Hz}$. $\lambda$ increases exponentially until the target rate is reached fairly early on in training, after which $\lambda$ gradually decreases. The slight increase of $\lambda$ near the end is caused by a drop in the learning rate (see appendix~\ref{apx:slowae-details}).}
\label{fig:lambda-and-aer}
\end{center}
\vskip -0.2in
\end{figure}

Concretely, at every training step $k$, we update $\lambda$ based on the AER estimated from the current batch of training data $\hat{R}_k$, and the target value $R_\mathrm{T}$:

\begin{equation}
    \lambda_{k + 1} = 
     \begin{cases}
     (1 + \delta) \lambda_k        & \text{if $\hat{R}_k > (1 + \epsilon) R_\mathrm{T}$ } \\
     (1 + \delta)^{-1} \lambda_k   & \text{if $\hat{R}_k < (1 + \epsilon)^{-1} R_\mathrm{T}$ } \\
     \lambda_k                     & \text{otherwise}
  \end{cases} ,
\end{equation}

where $\delta$ and $\epsilon$ are hyperparameters controlling the rate of change and the tolerance with respect to the rate estimate. Note that we update $\lambda$ multiplicatively, because meaningful values for this parameter can span many orders of magnitude. We use $\epsilon = 10^{-2}$ and $\delta = 10^{-3}$ in practice, and cap the value of $\lambda$ between $10^{-8}$ and $10^8$. Figure~\ref{fig:lambda-and-aer} shows the evolution of $\lambda$ and $\hat{R}_k$ over the course of SlowAE training.

\subsubsection{SlowAEs for speech}
\label{sec:slowaes-for-speech}

To learn representations of speech signals, we construct a SlowAE with an autoregressive decoder, using the group-sparse slowness penalty $\mathcal{L}_\mathrm{GS}^{\mathrm{slow}}$ and STQ. We use 4 channels and 15 quantisation levels ($k = 7$). With this setup, we can target AERs as low as $75 \mathrm{Hz}$ and still get reconstructions that are intelligible for the most part (see \S\ref{sec:experiments-aers}).

We represent waveforms as 8-bit mu-law companded values~\cite{wavenet}. Before companding, we find it helpful to inject a small amount of Gaussian noise at the decoder side: this helps obscure imperceptible details during relatively quiet portions of the signals (which mu-law companding would otherwise exacerbate), further reducing the event rate. The decoder is also conditioned on the speaker identity, which encourages the learnt representations to be speaker-independent~\cite{vqvae}. Following \citet{deltavae}, we enforce anti-causality in the encoder, so that the learnt representation captures information about the future of the signal at each point. Because the decoder is autoregressive, it already has access to past signal values. Anecdotally, we found that this restriction does not meaningfully affect the quality of the representations, and has a stabilising effect on training. Architecture and training details are provided in appendix~\ref{apx:slowae-details}.

\subsection{Run-length Transformers (RLTs)}

After training a SlowAE, we can convert speech signals to variable length event sequences, consisting of $(\mathrm{value}, \mathrm{length})$-tuples. We can model these with an autoregressive generative model, based on the Transformer architecture~\cite{transformer}. Transformers exclusively use attention mechanisms to combine information from different input elements, and do not encode any inductive biases beyond the importance of pairwise interactions between these elements. They are particularly well suited for modelling event sequences, because they make no assumptions about the underlying topology. This is unlike e.g. convolutional layers, which implicitly assume a fixed sampling interval between consecutive sequence elements (note that this specifically is not the case for event sequences).

In Transformers, any known topological structure can instead be modelled using input embeddings, such as the positional encoding used in the original Transformer model. We will make extensive use of this approach, which enables us to endow the model with knowledge about the underlying structure of the event sequences, reflecting the slowly varying representations before run-length encoding.

An important consideration for autoregressive models is the input ordering. Because SlowAEs will typically require multiple channels to enable faithful reconstruction of speech signals, the resulting slowly varying discrete representations are multi-channel. This causes some ambiguity: we could order the events by channel and then by time, or by time and then by channel. We choose the latter, as it yields a temporally ordered event sequence\footnote{\citet{spn} encounter a similar situation after separating images into multiple channels. They choose the `channels first' ordering, which is equally valid.}
For each event, we first predict the value and then the length.

\subsubsection{Multi-channel run-length encoding}

\begin{algorithm}[tb]
   \caption{Infer channels and offsets from a (partial) event sequence of length $N$ (with $C$ channels).}
   \label{algo:infer-channels-offsets}
\begin{algorithmic}
   \STATE {\bfseries Input:} sequence of events $(v_n, l_n)$.
   \STATE Initialize positions $p_c := 0$, $\forall c \in \{1, \ldots, C\}$
   \FOR{$i=1$ {\bfseries to} $N$}
   \STATE $c_i := \mathrm{argmin}_k (p_k)$
   \STATE $o_i := p_{c_i}$
   \STATE $p_{c_i} := p_{c_i} + l_i$
   \ENDFOR
   \STATE Return all channels $c_n$, offsets $o_n$.
\end{algorithmic}
\end{algorithm}

\begin{figure}[t]
\vskip 0.2in
\begin{center}
\centerline{\includegraphics[width=\columnwidth, trim=0 140 420 0, clip]{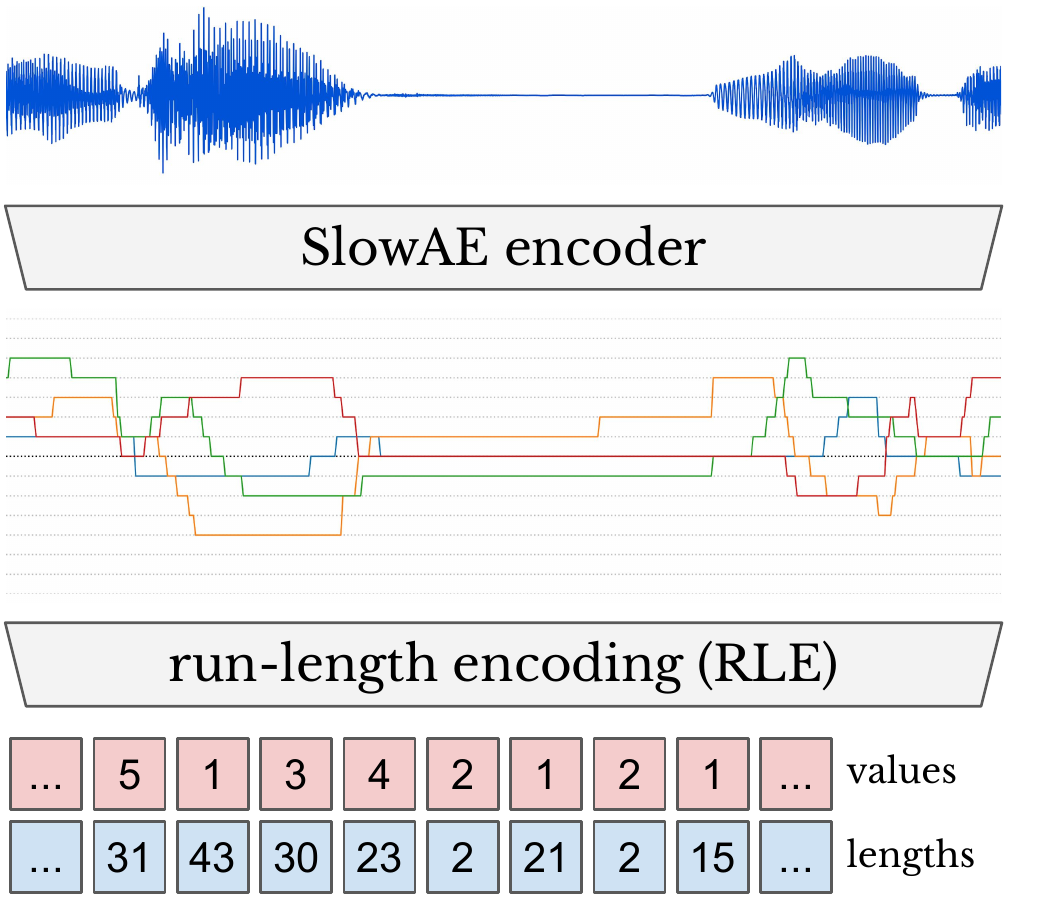}}
\caption{To extract event-based representations from speech signals, we first train a SlowAE and use its encoder to extract a multi-channel slowly varying discrete representation (middle). Next, we apply run-length encoding separately on each channel, and interleave the resulting sequences into a single event sequence of $(\mathrm{value},\mathrm{length})$-tuples (bottom).}
\label{fig:rle}
\end{center}
\vskip -0.2in
\end{figure}

We perform run-length encoding (RLE) on each channel of the slowly varying discrete representations, and \emph{interleave} the resulting $(\mathrm{value},\mathrm{length})$-tuples so they are ordered by time and then by channel index (see Figure~\ref{fig:rle}), resulting in a one-dimensional sequence of events. Given a fixed assignment of channel indices, this ordering is deterministic. As a result, we can recursively infer the channel and \emph{offset} (position in the representation before RLE) for each tuple, given only the sequence of event lengths so far (see Algorithm \ref{algo:infer-channels-offsets}).

For example, if we are given the interleaved event sequence $(2, 3), (0, 2), (1, 6), (3, 2), (4, 3)$, and we know that there are two channels, we can infer that the corresponding channel sequence is $1, 2, 2, 1, 1$, and the offset sequence is $0, 0, 2, 3, 5$. This decodes to $2, 2, 2, 3, 3, 4, 4, 4$ for channel 1, and $0, 0, 1, 1, 1, 1, 1, 1$ for channel 2.

Concretely, this means that the Transformer model can predict a single event at a time, and never has to predict its channel or offset, because they are uniquely determined by the preceding event sequence. Only the value and length need to be predicted. Note that we use a maximal run length of 256: runs longer than this are split up into multiple events.

\subsubsection{Architecture}
\label{sec:rlt-architecture}
\begin{figure}[t]
\vskip 0.2in
\begin{center}
\centerline{\includegraphics[width=\columnwidth, trim=0 60 420 0, clip]{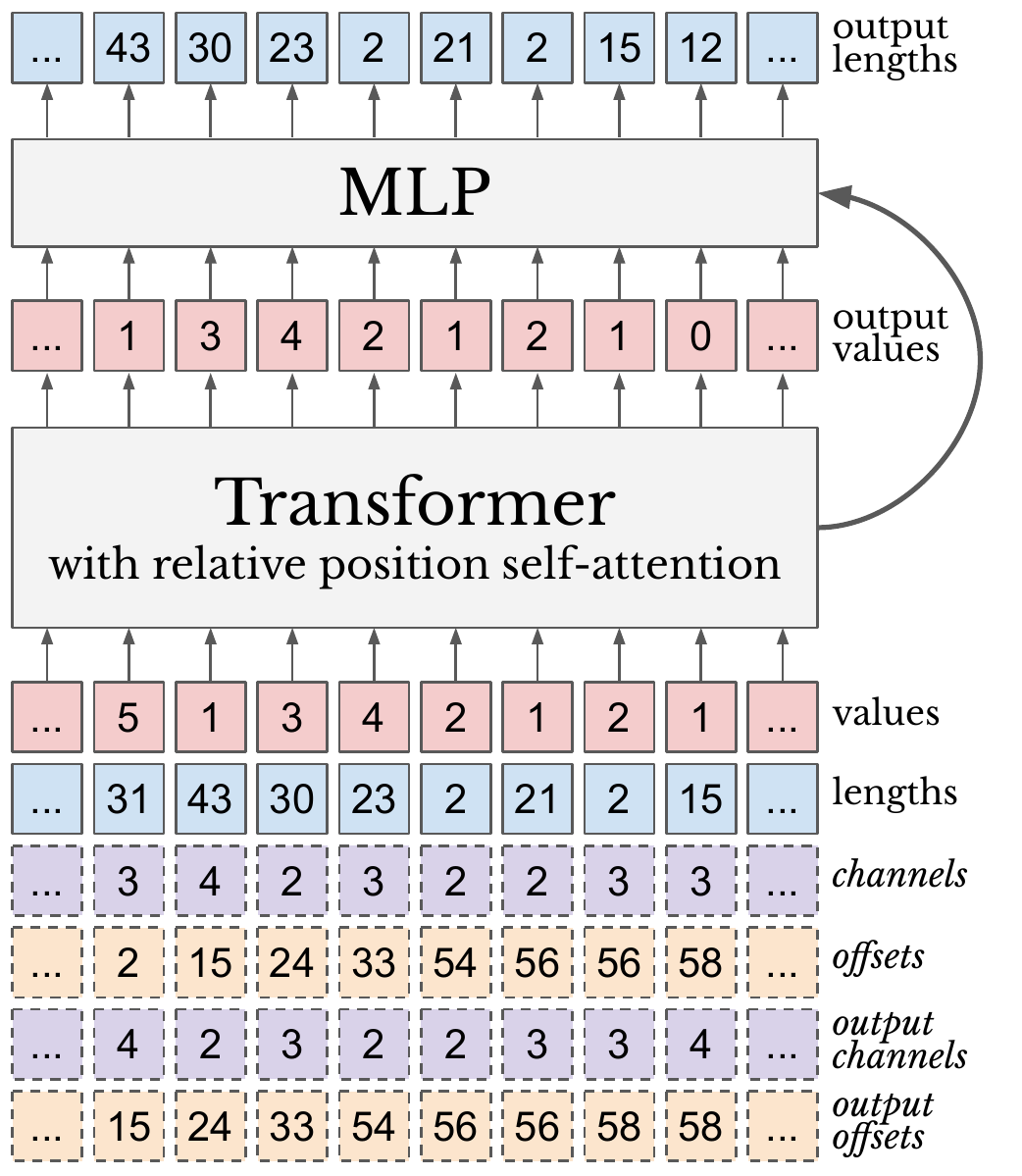}}
\caption{Schematic overview of the run-length Transformer (RLT) architecture. Input embeddings of $v_n$, $l_n$, $c_n$, $o_n$, $c_{n+1}$ and $o_{n+1}$ for each event are provided to a Transformer (the last four of these can be inferred from $l_1, \ldots, l_n$, see Algorithm~\ref{algo:infer-channels-offsets}). The next value $v_{n+1}$ is predicted from the Transformer output. An MLP which takes the Transformer output and embeddings of $v_{n+1}$ as input then predicts the next length $l_{n+1}$.}
\label{fig:transformer}
\end{center}
\vskip -0.2in
\end{figure}

The model architecture we use is shown schematically in Figure~\ref{fig:transformer}. The main component is a causally masked Transformer with relative position self-attention, which contains the bulk of the learnable parameters. The input consists of a set of learnable embeddings which are additively combined, encoding the current value ($v_n$), length ($l_n$), channel ($c_n$) and offset ($o_n$), as well as the output channel ($c_{n+1}$) and offset ($o_{n+1}$). Although these last four are deterministic functions of $l_1, \ldots, l_n$ (provided that the number of channels $C$ is known), we find that explicitly providing this information in the form of input embeddings dramatically accelerates convergence (see \S\ref{sec:experiments-input-embeddings}).

We observe that the relationship between the length of an event and its value, conditioned on all preceding events, is fairly straightforward to model. Empirically, we find that it suffices to stack a two-layer multi-layer perceptron (MLP) on top of the Transformer and feed it an embedding of $v_{n+1}$ as a side-input to predict $l_{n+1}$. As a result, we do not need to flatten the $(\mathrm{value},\mathrm{length})$-tuples into an alternating sequence. This is significant, as the cost of the attention mechanisms inside the Transformer is quadratic in the sequence length. Architecture and training details are provided in appendix~\ref{apx:rlt-details}.

To sample from the model, we alternate between sampling $v_n$ and $l_n$, inferring other inputs on the fly using Algorithm~\ref{algo:infer-channels-offsets} when we move to the next event in the sequence. The resulting event sequence can then be run-length decoded, and rendered to an audio waveform using the SlowAE decoder.

\section{Language modelling in the speech domain}
\label{sec:language-modelling-in-the-speech-domain}

Language models traditionally operate on textual representations, but the written form is not the only way in which language manifests itself. In fact, humans usually acquire language first by listening and speaking, before learning to read and write (if at all). While the spoken and written form of a language are closely related, there are important differences between both modalities: spoken language often diverges significantly from written language through dialects, speech registers or historical orthographies. It also carries additional meaning through intonation, tempo, rhythm, stress and volume (reflecting e.g. emphasis, evidentiality or emotion). It tends to have a more subjective quality than its written counterpart, and it is often more context-dependent as well.

We therefore want to explore language modelling in the speech domain, in an attempt to capture these different aspects of language beyond grammar and vocabulary. We also want to demonstrate the possibility for a model to learn about grammar and vocabulary from speech only -- an ability that humans also have. Finally, speech language models can also be applied to languages with limited written resources, including languages that have no written form at all~\cite{ethnologue}.

The information content of speech signals is orders of magnitude larger than that of the corresponding textual transcriptions, so we cannot simply apply state-of-the-art large-scale language models~\cite{gpt3} directly to raw waveforms. Instead, we use SlowAEs (\S\ref{sec:slowaes}) for representation learning, because they capture enough high-level information to reconstruct intelligible speech signals, while also making abstraction of irrelevant details (see Table~\ref{tab:bps}). This makes the task of speech-level language modelling tractable, albeit still significantly more computationally expensive than its textual analogue. While we have only tested this method on English speech, nothing about it is specific to European languages (or to English in particular), and we do not encounter the incompatibility issues that textual language models may face in this context (e.g. other writing systems requiring different tokenisation strategies).

We note and acknowledge that our use of a dataset of audiobooks for this work (see \S\ref{sec:datasets}), while practical, comes with important limitations: audiobooks contain written language that is read aloud, and thus represent only one particular restricted mode of speech, which is relatively clean compared to other modes. Extending our work to datasets consisting of e.g. conversational speech is an important line of future work, which will undoubtedly present additional challenges.

\section{Related work}
\label{sec:related-work}

\paragraph{Slowness and predictability priors.} Since the introduction of slow feature analysis~\cite{sfa,mlsfa}, slowness as an inductive bias in machine learning models has been studied extensively. Most relevant to our work is its application to variational autoencoders (VAEs, \citet{vae1,vae2}) to prevent posterior collapse~\cite{posterior-collapse} by \citet{deltavae}, and to learn disentangled representations by \citet{slowvae}. We refer to the latter for a detailed overview of the application of L1 and L2 slowness priors in other domains. Enforcing a learnt representation to be \emph{predictable} can be seen as a relaxed form of slowness. This is the inductive bias used in contrastive predictive coding (CPC, \citet{cpc}).

\paragraph{Structured discrete representation learning.} \citet{somvae} equip VQ-VAE~\cite{vqvae} with a Markov model in latent space, which is trained jointly with the autoencoder, to improve interpretability. \citet{vqcpc} combine CPC with a VQ bottleneck to learn representations for symbolic music generation. \citet{discretalk} use VQ-VAE to learn discrete representations of speech waveforms, and compress them using byte-pair encoding (BPE, \citet{bpe}), casting text-to-speech as a machine translation problem. \citet{chorowski-segments,chorowski-slowvq,convdmm} use VQ-VAE to learn speech representations, with various regularisation strategies to make them reflect phonetic structure. \citet{stdp} use spiking neural nets for representation learning.

\paragraph{Structure discovery.}
Learning discrete representations is only one possible route to achieve unsupervised structure discovery. \citet{blindseg} analyse frame-wise predictions of speech features to determine phoneme boundaries. \citet{chunkers} suggests a similar approach for dividing sequences into `chunks' based on compressibility. \citet{iic} use mutual information maximisation for unsupervised image classification and segmentation. \citet{diffseg} propose a differentiable relaxation to learn segmentations of sequences. In the visual domain, several object-centric representation learning models which discover objects without supervision have been proposed recently~\cite{monet,iodine,slotattention,visualtransformers}.

\paragraph{Variable-sized latents.} While we consider representations with variable rates along the sequence dimension, one can also vary the capacity of latent representations along other axes. This is especially relevant when they do not have any topological structure. Examples of such approaches are the Indian buffet process~\cite{indianbuffet}, Stick-breaking VAEs~\cite{stickbreaking}, Infinite restricted Boltzmann machines~\cite{infinite-rbms} and nested dropout~\cite{nested-dropout}.

\paragraph{`Language-ification'.} Recent advances in large-scale textual language modelling have inspired many researchers to try and fit other tasks into the same mold, by making the input and output representations resemble token sequences. Besides the work by \citet{discretalk} mentioned earlier, this approach has been used for speech-to-speech translation~\cite{uwspeech}, video representation learning~\cite{videobert}, music generation~\cite{jukebox} and image generation~\cite{dalle}.

\paragraph{Event-based representations.} In symbolic music modelling, event-based representations are a natural fit as the popular MIDI\footnote{Musical Instrument Digital Interface} format is event-based, yet they were not very commonly used until recently. They have significantly advanced the state of the art since~\cite{perfrnn,musictransformer,musenet,feeling}. Event-based representations of visual data can be obtained from \emph{event cameras}, and \citet{eventcameras} provide a survey of recent machine learning approaches in this domain.

\paragraph{Asynchronous models.} By and large, machine learning models of perceptual data assume that the input has a topological structure with a fixed quantum, i.e. the spacing between pixels in a grid or sampling points in a temporal signal is constant. However, a rich body of literature describes \emph{asynchronous} models that forgo this assumption (as run-length Transformers do). Many such variants of recurrent and convolutional neural networks have been explored~\cite{discrete-event-rnns,phasedlstm,arcnns,asyncrnns,latentodes,intprednets}. Other models operate on fixed-quantum inputs, but adaptively allocate capacity\footnote{This idea is also closely related to \emph{conditional computation}, for which \citet{surprisal} provide a comprehensive literature review.}~\cite{air,act,glimpses,skiprnn,spatialact,hardattention,effseg,pointrend,surprisal}. Run-length Transformers do this implicitly because they use more computation for parts of speech signals which require more events to represent. Some models operate directly on compressed representations to achieve a similar effect~\cite{dctransformer}, though usually some topological structure is preserved~\cite{jpeg-nets,compressed-video}.

\paragraph{Speech language models.} In concurrent work, \citet{spokenlm} develop language models in the speech domain, and evaluate them for speech resynthesis and generating novel utterances. They describe three different strategies for learning continuous intermediate representations, and use a post-hoc quantisation strategy to make them suitable for generative model training. They also introduce several evaluation metrics based on human ratings and text transcriptions obtained from a speech recognition model.

\section{Experiments and results}
\label{sec:experiments-and-results}

In this section, we compare different variants of SlowAEs (\S\ref{sec:experiments-slowaes}) and RLTs (\S\ref{sec:experiments-rlts}), and show how we can quantitatively evaluate the former using an auxiliary speech recognition model (\S\ref{sec:eval-through-asr}). We also investigate several approaches to improve sample quality (\S\ref{sec:sampling-and-reranking}).

\subsection{Datasets}
\label{sec:datasets}

We use the \textbf{Libri-light} dataset~\cite{librilight}, which consists of roughly 60,000 hours ($\approx$ 7 years) of speech at $16 \mathrm{kHz}$, obtained from the LibriVox\footnote{\url{https://librivox.org/}} repository of free public domain audiobooks. Recordings from various speakers of about 10,000 books are included. Assuming a speaking rate of 60 words per minute\footnote{This is a very conservative estimate, the true rate is probably well over 100 words per minute.}, we estimate it to contain approximately 216M words. This is a comparable order of magnitude to the WikiText-103 dataset (103M tokens, \citet{wikitext103}), which has been used extensively to train textual language models. It should therefore be large enough to enable models to learn about language-level structure in speech signals.

We also make use of the much smaller \textbf{LibriTTS} dataset~\cite{libritts}, specifically to train SlowAE models. This dataset is also sourced from LibriVox, but it contains only 585 hours of speech, which is of a much higher quality on average. We downsample the audio from $24 \mathrm{kHz}$ to $16 \mathrm{kHz}$ to match Libri-light.

While the Libri-light dataset is primarily intended for unsupervised pre-training of speech recognition models, we use it for generative modelling instead. We believe this is acceptable
as the audio recordings as well as the books themselves are all in the public domain. However, we are conscious that our models are not just reproducing language contained in the books, but may also imitate the voices of the speakers. Although LibriVox recordings have previously been used in this manner (e.g. for text-to-speech research as part of the LibriTTS dataset), we have made the deliberate choice not to render any samples from generative models using voices from these two datasets.

Instead, we train separate `post-hoc' decoders using a \textbf{set of proprietary recordings} from a single speaker, which are explicitly intended to train speech synthesis models (see appendix \ref{apx:post-hoc-details} for training and architecture details). Using separate decoders for representation learning and reconstruction~\cite{ham} also enables us to generate $24 \mathrm{kHz}$ audio, even though the SlowAE and RLT models were exclusively trained on $16 \mathrm{kHz}$ audio, and to remove the noise injection step (which improves reconstruction quality, see \S\ref{sec:slowaes-for-speech}). We also have temporally aligned phoneme-level annotations for these recordings (at $200 \mathrm{Hz}$), which we use to train simple speech recognition models for evaluation purposes (\S\ref{sec:eval-through-asr}).

\subsection{Slow autoencoders}
\label{sec:experiments-slowaes}
We have made quite a few design choices to arrive at the slow autoencoder model as described earlier (\S\ref{sec:slowaes}), which we use to extract representations for generative modelling. It has 4 channels and 15 quantisation levels, uses Schmitt trigger quantisation, and is trained with a group-sparse slowness penalty $\mathcal{L}^{slow}_{GS}$ with an AER target of $75 \mathrm{Hz}$. We will refer to this as our \emph{reference model}, and ablate these choices individually in the next few subsections. Reconstructions from the various model variants we describe can be listened to at \url{https://vdrl.github.io/}.

\subsubsection{Evaluation through speech recognition}
\label{sec:eval-through-asr}
To evaluate the representations, we have relied quite strongly on subjective assessments of the intelligibility of the reconstructions, as constructing a quantitative metric for this is challenging. We have attempted to do this using a speech recognition (ASR) model, but we note that the resulting measurements we report should be used as a guide only, and do not always correlate with our own subjective comparisons.

Concretely, we reuse the set of proprietary recordings which we also used for post-hoc decoder training, to train a convolutional ASR model that predicts aligned phoneme labels from spectrogram frames. We note that this is a fairly simplistic approach compared to the state of the art in ASR, where aligned phoneme labels generally are not required~\cite{ctc}. In our setting however, we are not interested in obtaining state-of-the-art speech recognition results, but rather a metric that is a good proxy for intelligibility. We use the phoneme accuracy on a holdout set of 16 4-second utterances for this purpose. Details about the ASR model can be found in appendix~\ref{apx:asr-details}.

\subsubsection{Average event rates}
\label{sec:experiments-aers}
\begin{figure}[t]
\vskip 0.2in
\begin{center}
\centerline{\includegraphics[width=\columnwidth]{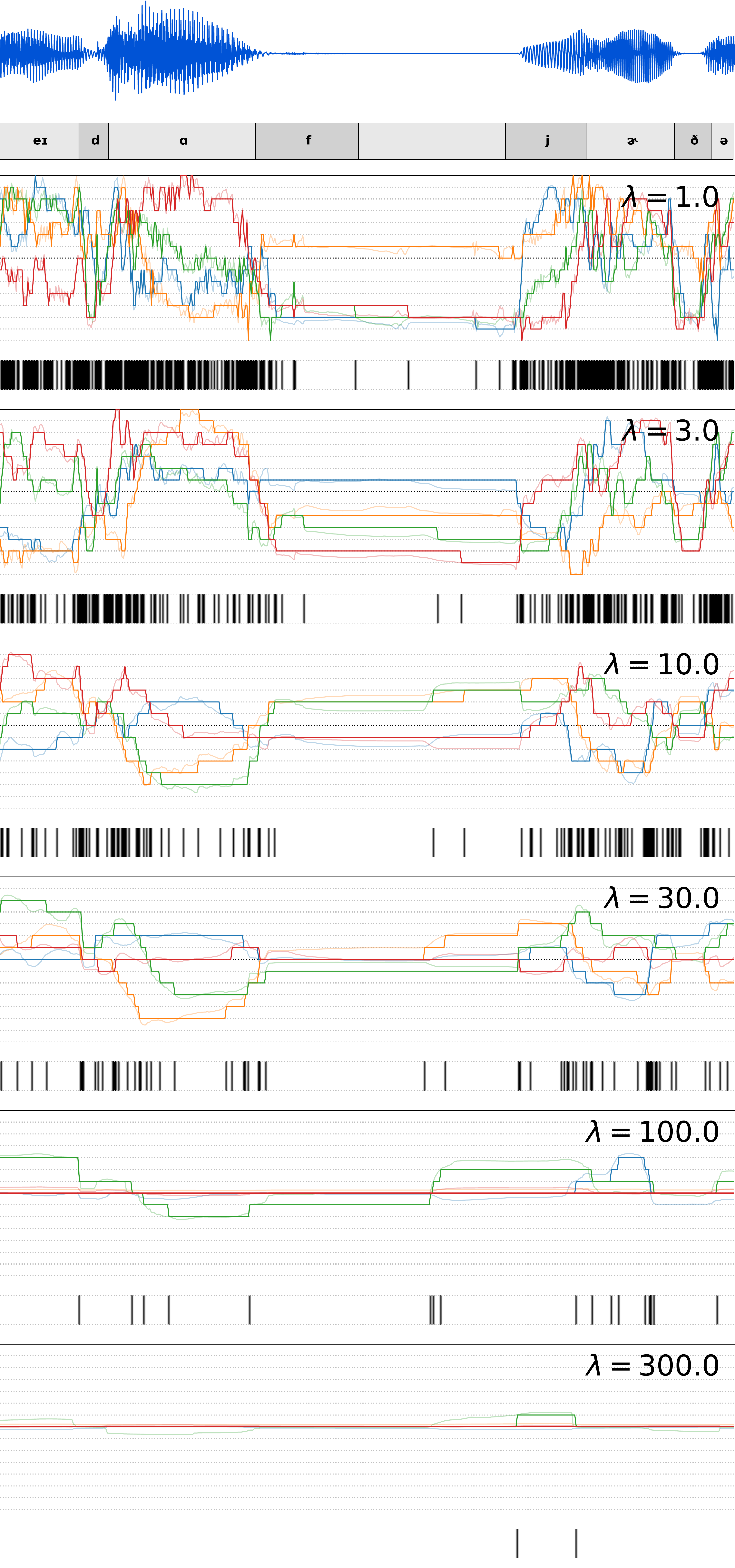}}
\caption{Slow discrete representations $\mathbf{z'}$ for a one-second speech excerpt, produced by SlowAEs with different (fixed) slowness penalty weights (see \S\ref{sec:experiments-aers}). Faint lines show the underlying continuous output $\mathbf{z}$. `Barcode' plots show event density. Event density varies over time for all values of $\lambda$, as shown by the barcode plots. Event rates vary significantly even for values of $\lambda$ that are less than an order of magnitude apart, making manual tuning challenging.}
\label{fig:compare-lambdas}
\end{center}
\vskip -0.2in
\end{figure}

\begin{figure}[t]
\vskip 0.2in
\begin{center}
\centerline{\includegraphics[width=\columnwidth]{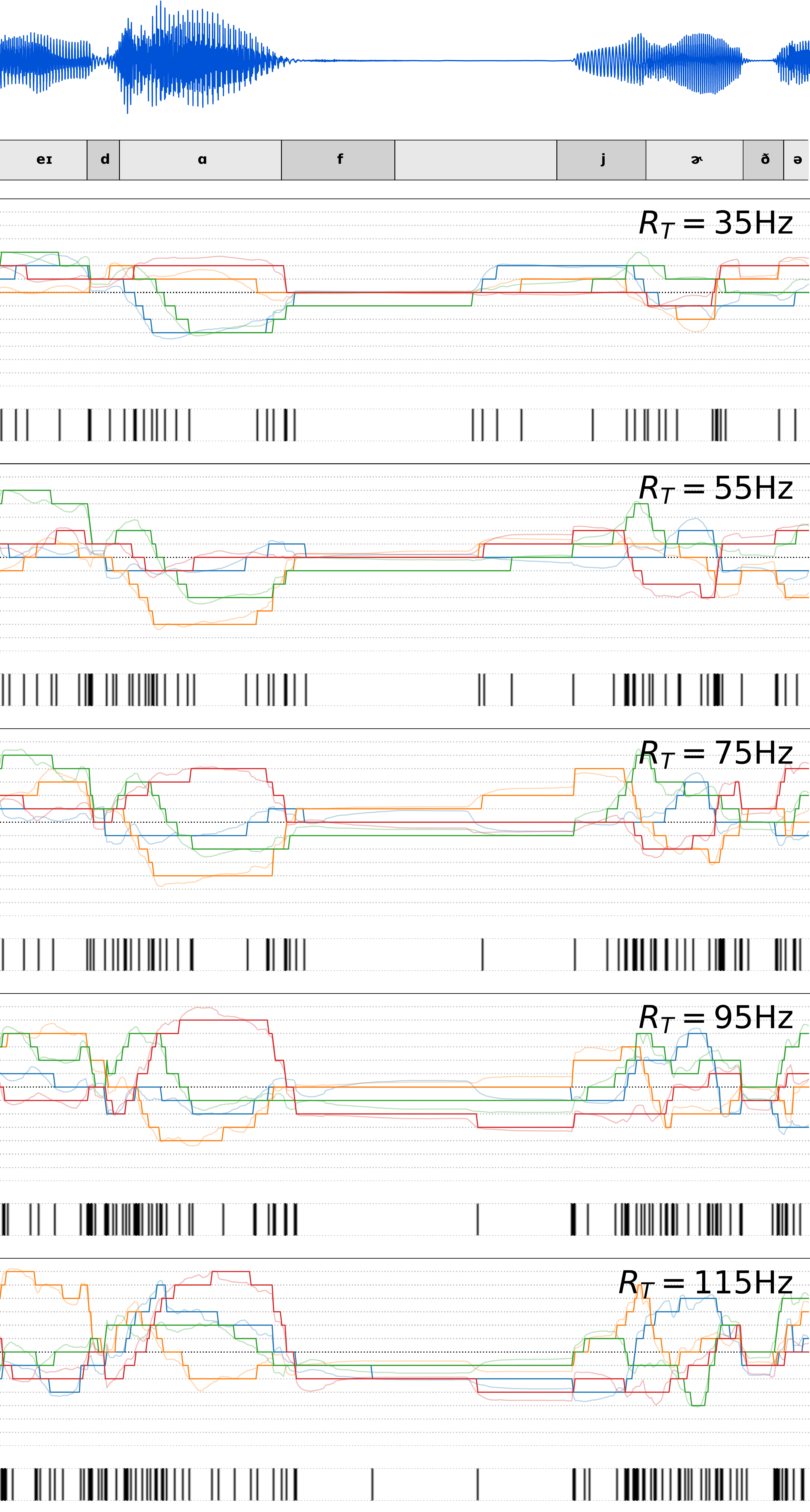}}
\caption{Slow discrete representations $\mathbf{z'}$ for a one-second speech excerpt, produced by SlowAEs with automatic slowness penalty weighting for different targets average event rates (AERs) $R_\mathrm{T}$ (see \S\ref{sec:automatic-penalty-weighting} and \S\ref{sec:experiments-aers}). Faint lines show the underlying continuous output $\mathbf{z}$. `Barcode' plots show event density. Event density varies over time in a similar way across all the displayed rates, as shown by the barcode plots. Higher targets result in a better usage of the quantisation levels.}
\label{fig:compare-rates}
\end{center}
\vskip -0.2in
\end{figure}

We study the effect of the slowness penalty weight $\lambda$ by training several models with fixed weight values, as well as several models with automatic penalty weighting and different average event rate (AER) targets. The resulting representations are visualised in Figures~\ref{fig:compare-lambdas} and \ref{fig:compare-rates} respectively. The sensitivity of the AER to $\lambda$ is clear, and automatic penalty weighting is an effective approach to control it.

\begin{figure*}[t]
\vskip 0.2in
\begin{center}
\centerline{\includegraphics[width=\textwidth]{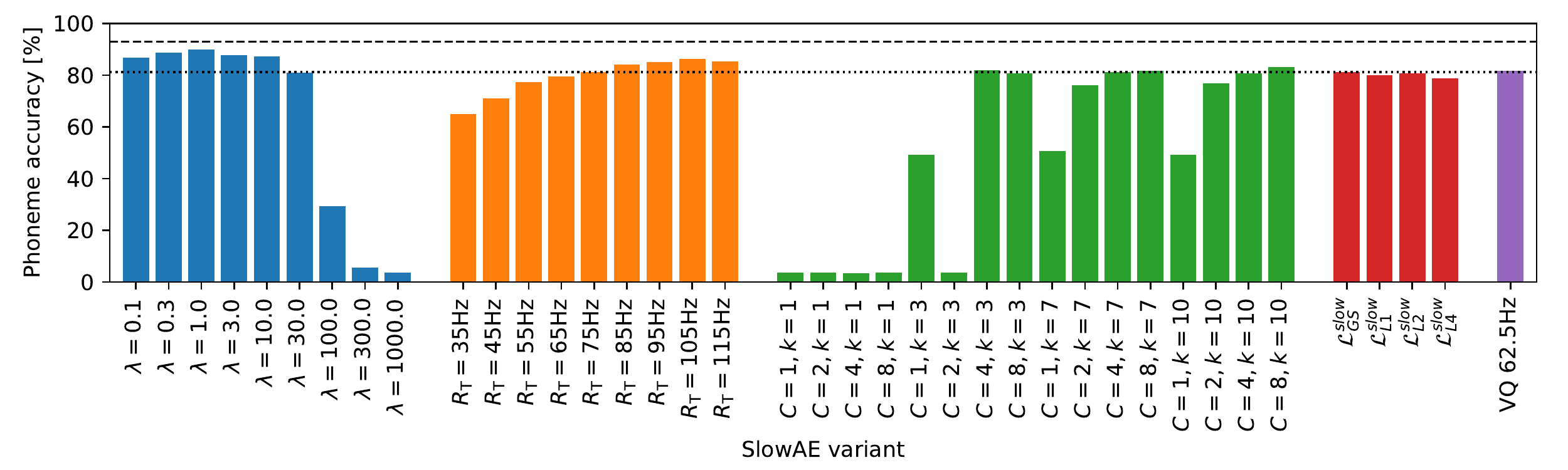}}
\caption{Phoneme accuracies of a speech recognition model on reconstructions from various SlowAE variants. The dashed line indicates the accuracy obtained on the original recordings. The dotted line indicates the accuracy of the reference model that we use for further experiments ($C=4,k=7,R_\mathrm{T}=75\mathrm{Hz},\mathcal{L}^{slow}_{GS}$). From left to right: different values of the slowness penalty weight $\lambda$ ({\color{blue} blue}); different average phoneme rate (AER) targets, $\lambda$ is automatically adapted ({\color{orange} orange}); different numbers of channels ($C$) and quantisation levels ($2k + 1$), with $R_\mathrm{T} = 75 \mathrm{Hz}$ ({\color{green} green}); different types of slowness penalty ({\color{red} red}); vector quantisation baseline ({\color{violet} purple}).}
\label{fig:phoneme-accuracies}
\end{center}
\vskip -0.2in
\end{figure*}

To quantify the intelligibility of these representations, we use the post-hoc decoders to render them to audio and then predict phoneme sequences using an ASR model (\S\ref{sec:eval-through-asr}). We compute the accuracies when compared to the ground truth phoneme sequences and report them in Figure~\ref{fig:phoneme-accuracies}.

First, we note that the accuracy starts dropping off quite suddenly and dramatically for $\lambda \geq 30$, which demonstrates the difficulty of manual tuning. Using automatic slowness penalty weighting, we find that the accuracy increases quite smoothly as we increase the target AER $R_\mathrm{T}$. Note that the value we use for our reference model ($R_\mathrm{T} = 75 \mathrm{Hz}$) does not have the highest accuracy, but it is high enough for the reconstructions to be intelligible most of the time. We choose this lower target value because it reduces the length of the resulting event sequences after run-length encoding, which will in turn make the generative modelling task easier.

\begin{figure}[t]
\vskip 0.2in
\begin{center}
\centerline{\includegraphics[width=\columnwidth]{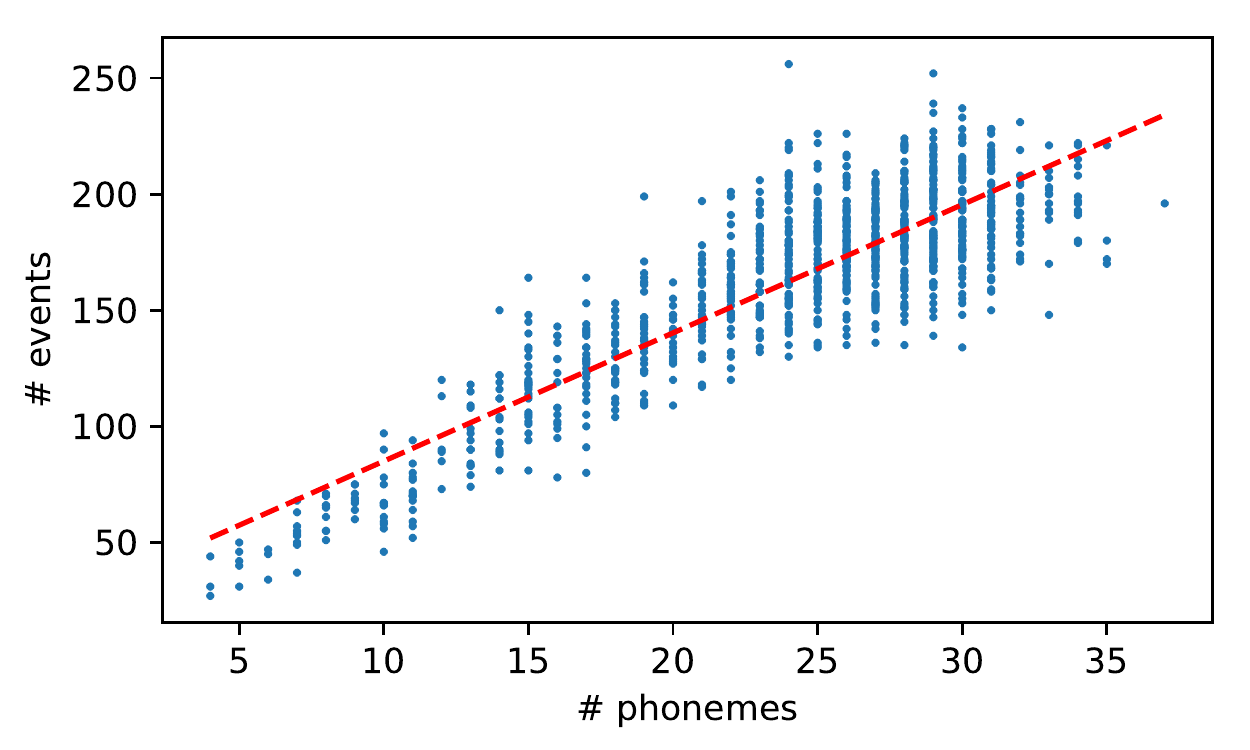}}
\caption{Scatter plot of event sequence lengths vs. number of phonemes for 1024 two-second excerpts from the proprietary single-speaker dataset. There is a strong correlation despite the fact that the representation learning procedure is entirely unsupervised (Pearson's $\rho = 0.866$, Spearman's $\rho=0.753$).}
\label{fig:scatter-phonemes-events}
\end{center}
\vskip -0.2in
\end{figure}

In practice, we find that SlowAEs do not perfectly preserve prosody, even at high AERs, but they do tend to capture aspects of speech beyond phoneme identity to a reasonable extent. The event rate is also strongly correlated with the phoneme rate. We demonstrate this by using our reference model to infer event sequences for 1024 two-second excerpts from the proprietary single-speaker dataset. Figure~\ref{fig:scatter-phonemes-events} shows a scatter plot of the number of phonemes vs. the number of events in these excerpts, as well as a line of best fit (Pearson's $\rho = 0.866$, Spearman's $\rho=0.753$). While the events in these sequences may not be interpretable on an individual basis, they clearly mirror the density of salient information in the speech signals quite accurately.

\subsubsection{Number of channels, quantisation levels}
\label{sec:num-channels-levels}
When varying the number of channels and quantisation levels while keeping the AER fixed ($R_\mathrm{T} = 75 \mathrm{Hz}$), phoneme accuracies can be significantly impacted, as shown in Figure~\ref{fig:phoneme-accuracies} (3rd group). A single channel is clearly insufficient, and 4 is better than 2, but the difference between 4 and 8 channels is less clear-cut. For the number of quantisation levels, $k = 10$ (21 levels) provides no clear benefit over $k = 7$ (15 levels).

\begin{figure}[t]
\vskip 0.2in
\begin{center}
\centerline{\includegraphics[width=\columnwidth]{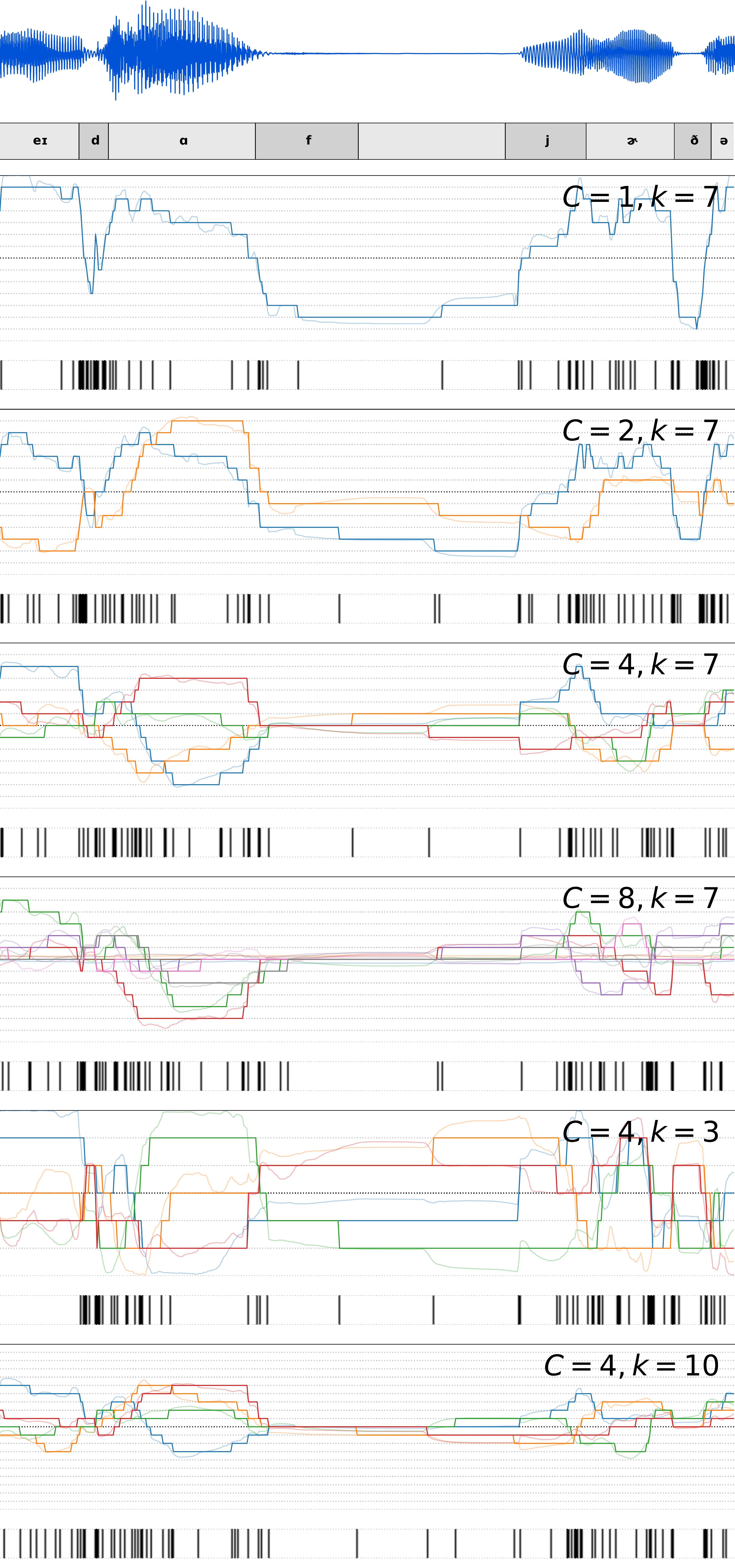}}
\caption{Slow discrete representations $\mathbf{z'}$ for a one-second speech excerpt, produced by SlowAEs with different numbers of channels ($C$) and quantisation levels ($2k + 1$). $R_\mathrm{T}$ is fixed at $75 \mathrm{Hz}$. Faint lines show the underlying continuous output $\mathbf{z}$. `Barcode' plots show event density. The first four variants differ only in the number of channels. A higher number of channels makes the representation more expressive, but requires each individual channel to move more slowly to hit the target average event rate (AER). The last two variants have four channels, but differ in the number of quantisation levels. $k=3$ results in a fairly large quantisation error, while $k=10$ results in inefficient use of the quantisation bins.}
\label{fig:compare-channels-levels}
\end{center}
\vskip -0.2in
\end{figure}

When we visually inspect the learnt representations (shown in Figure~\ref{fig:compare-channels-levels}, it becomes clear that using a high number of channels or levels results in less efficient usage of the capacity of the bottleneck. $C=4,k=7$ strikes a nice balance between expressivity and efficiency.

\subsubsection{Slowness penalties}
\label{sec:experiments-slowness-penalties}
The different types of slowness penalty discussed in \S\ref{sec:slowness-penalty} affect the appearance of the learnt discrete representations quite significantly, as shown in Figure~\ref{fig:compare-codes}. In terms of phoneme accuracy, they all perform comparably (Figure~\ref{fig:phoneme-accuracies}), but this does not reflect our subjective assessment, which slightly favours the group-sparse penalty over the other variants. We recommend that the reader listen to the provided reconstructions if they wish to verify this.

\begin{figure*}[t]
\vskip 0.2in
\begin{center}
\centerline{\includegraphics[width=\textwidth]{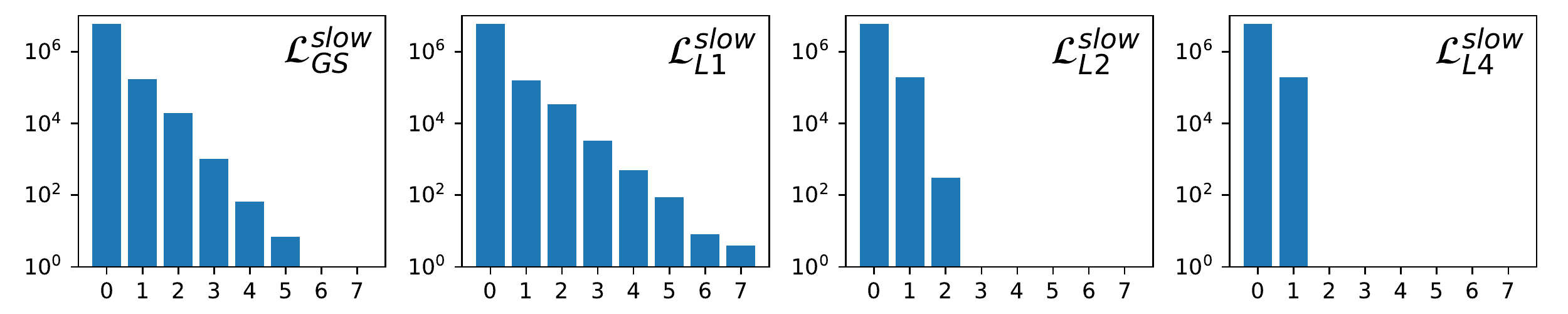}}
\caption{Histograms of `jumps', $|z'_{t+1,c} - z'_{t,c}|$, estimated on a set of 1024 3-second clips from our proprietary single-speaker dataset, for SlowAEs with different slowness penalties. Note the logarithmic y-axis. Sparse and group-sparse slowness penalties allow larger jumps.}
\label{fig:jump-plots}
\end{center}
\vskip -0.2in
\end{figure*}

Figure~\ref{fig:jump-plots} shows a histogram of the absolute difference in value between two subsequent events in the same channel $|z'_{t+1,c} - z'_{t,c}|$ (`jumps'), for different slowness penalties, measured on a set of clips from our proprietary single-speaker dataset (which the encoder models were not trained on). This clearly demonstrates that sparse and group-sparse slowness penalties make the discrete representations more flexible, by allowing for larger sudden changes in value.

\subsubsection{Schmitt trigger quantisation, noise}
\begin{figure}[t]
\vskip 0.2in
\begin{center}
\centerline{\includegraphics[width=\columnwidth]{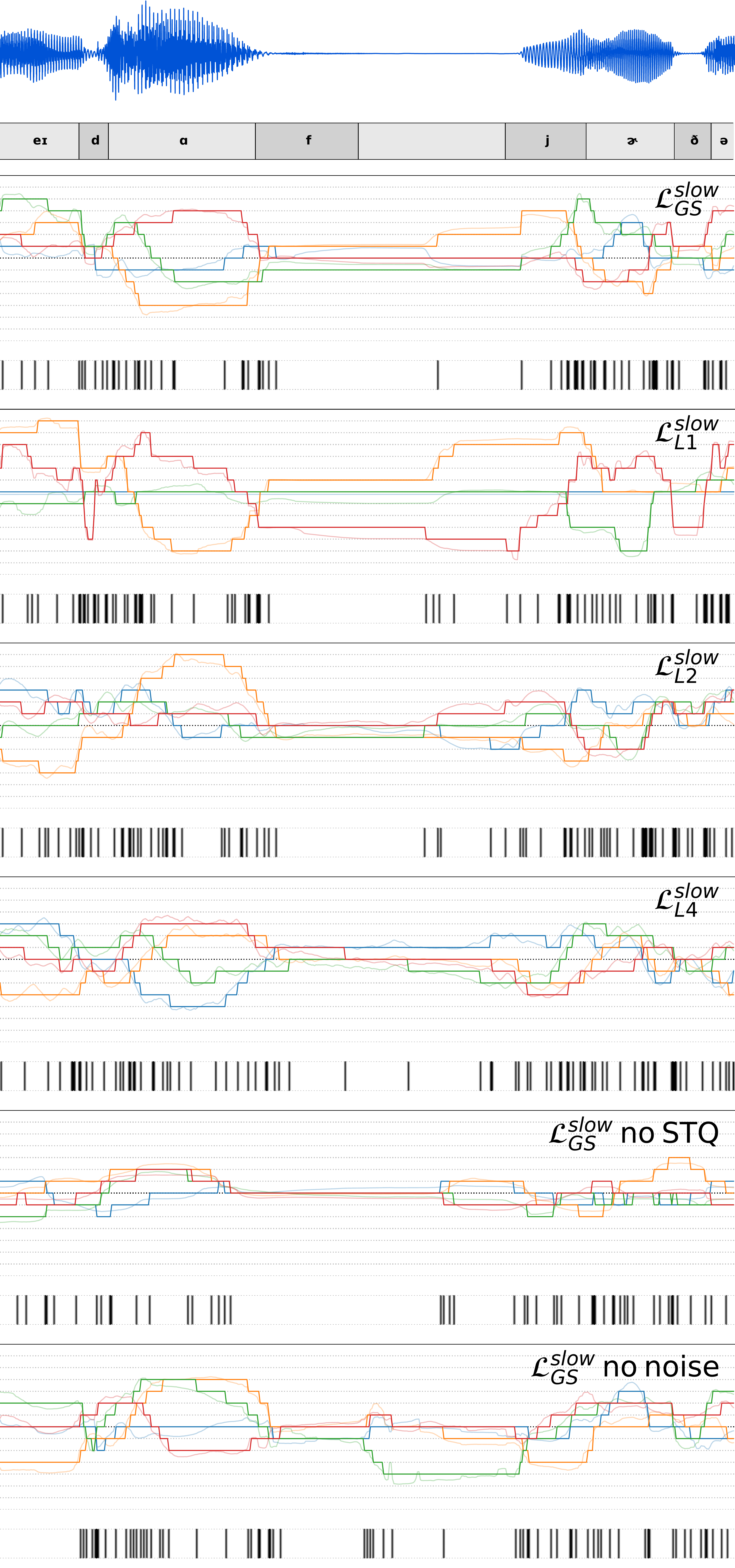}}
\caption{Slow discrete representations $\mathbf{z'}$ for a one-second speech excerpt, produced by different SlowAE variants. Faint lines show the underlying continuous output $\mathbf{z}$. `Barcode' plots show event density. From top to bottom: SlowAE with \textbf{group-sparse slowness}; with \textbf{L1 slowness} (some channels collapse to zero); with \textbf{L2 slowness} (values change more gradually, rather than in bursts); with \textbf{L4 slowness} (values change even more gradually); with group-sparse slowness but \textbf{no Schmitt-trigger quantisation} (inefficient use of quantisation levels); with group-sparse slowness but \textbf{no noise injection} at the decoder side (capacity is wasted in silent parts of the input signal).}
\label{fig:compare-codes}
\end{center}
\vskip -0.2in
\end{figure}

The importance of Schmitt trigger quantisation (\S\ref{sec:stq}) and noise injection at the decoder side (\S\ref{sec:slowaes-for-speech}) are also demonstrated in Figure~\ref{fig:compare-codes} (bottom): ablating the former results in much less efficient use of quantisation bins, whereas ablating the latter results in more events being generated for silent parts of the input signal.

\subsection{Run-length Transformers}
\label{sec:experiments-rlts}
To investigate and compare RLT architectures, we train several unconditional models with approximately 1 billion parameters (1M updates with sequences of 512 events, batch size 256). We also train a final model with about 2.4 billion parameters with 8 times more data (1M updates with batch size 2048), conditioned on book identity embeddings.

\subsubsection{Input embeddings}
\label{sec:experiments-input-embeddings}
We experiment with different combinations of input embeddings. In theory, only $v_k, l_k, k = 1, \ldots, n$ are necessary to be able to predict $v_{n+1}$ without making any independence assumptions. However, as mentioned in \S\ref{sec:rlt-architecture}, we find that it is helpful to provide additional derived information about the underlying structure of the sequence: channels and offsets $c_k, o_k, k = 1, \ldots, n + 1$, which can be inferred from the lengths $l_k, k = 1, \ldots, n$. We ablate this by training RLT models with various subsets of these embeddings, and also experiment with absolute positional embeddings.

\begin{figure*}[t]
\vskip 0.2in
\begin{center}
\centerline{\includegraphics[width=\textwidth]{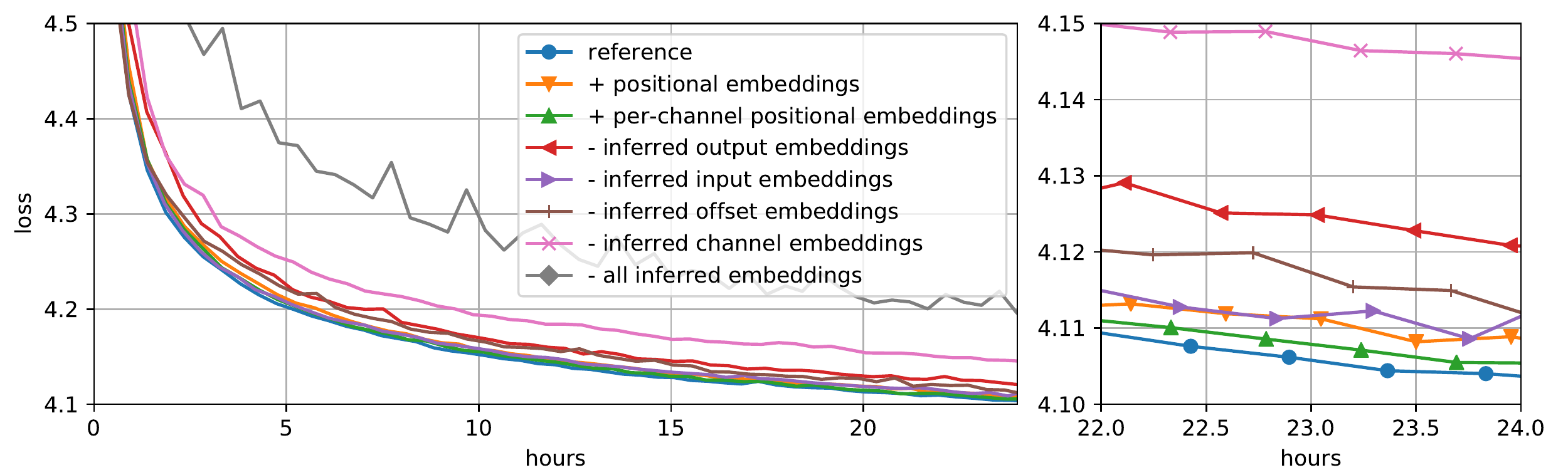}}
\caption{Negative log-likehood on a holdout set during run-length Transformer training, for various variants with different input embeddings. Adding absolute positional embeddings or removing any of the channel or offset embeddings slows down convergence. On the right, we show a zoomed in version of the same curves with markers, so the different ablations can be distinguished more clearly.}
\label{fig:input-embeddings}
\end{center}
\vskip -0.2in
\end{figure*}

Figure~\ref{fig:input-embeddings} shows the negative log-likelihood on a holdout set (from the LibriTTS dataset) during 24 hours of training. We compare the reference model (with channel and offset embeddings) to a number of ablations. The importance of channel and offset embeddings is definitively shown, with much worse convergence when either (or both) are removed. Embeddings of the output channel $c_{n+1}$ and offset $o_{n+1}$ are more important than the corresponding input embeddings $c_n$ and $o_n$, but both help to improve results. We also try two variants of embeddings of the absolute position in the event sequence: global position embeddings and per-channel embeddings. Both of these seem to make the results slightly worse. This is not too surprising, as we make use of relative position self-attention in the Transformer. Note that the model performs much worse and training becomes somewhat less stable when all inferred embeddings are removed.

\subsubsection{Sampling and reranking}
\label{sec:sampling-and-reranking}
To sample from trained RLTs, we can recursively make predictions for values and lengths (in an alternating fashion) and obtain event sequences. These can then be rendered to audio using post-hoc decoders (see \S\ref{sec:datasets}). Various audio samples from the models described in this section are available at \url{https://vdrl.github.io/}.

We investigate two methods for controlling the model output to some degree: \emph{book identity conditioning} and \emph{completion sampling}. First, when training an RLT on an event sequence, we can condition the model on the identity of the corresponding audiobook. We do this by adding an additional input embedding, which is broadcast across the event sequence. This then allows us to draw samples that resemble the content of a specific book at test time. To also enable unconditional sampling with the same model, we randomly replace the book identity embedding by a `catch-all' unconditional embedding for 10\% of training examples.

Second, we can present the model with an event sequence obtained from an existing recording, and ask the model to complete it. This allows us to prime the model to generate speech about a particular topic, and with a particular prosody. We can then assess to what extent the model captures the grammar and semantics of the priming sequences. The same approach has recently been used to make very large textual language models perform various tasks without any further training or finetuning~\cite{gpt3}. 

In practice, we find that a significant number of samples are unnatural or unintelligible. In autoregressive models, this commonly occurs because we recursively predict outputs based on previous predictions, whereas the model was only trained to do next-step prediction based on ground truth. This means that it is not robust to prediction errors, and they will tend to accumulate as a result (exposure bias, \citet{expbias1,expbias2}).

To improve sample quality, we employ two different strategies: \emph{nucleus sampling} and \emph{intelligibility reranking}. First, we modify the distributions from which event values and lengths are sampled to account for the fact that the model cannot predict low probability events as accurately as high probability events. We use nucleus sampling~\cite{nucleus}, which truncates the distributions by removing the least likely values which account for $1 - p$ of the probability mass, and then renormalising them. We use $p=0.8$ for both values and lengths for all samples, which noticeably improves overall sample quality.

Second, we repurpose the ASR model we trained to measure the intelligibility of SlowAE reconstructions (\S\ref{sec:eval-through-asr}) to measure the intelligibility of generated samples instead. We do not have aligned ground truth phoneme information for these samples, but we can measure the average entropy of the phoneme predictions as a proxy for model confidence. If we draw a large number of samples, we can then rank them according to this entropy. We find that samples with low phoneme prediction entropy are usually of higher quality. Similar reranking approaches have previously been used to select high-quality samples from generative models of images~\cite{vqvae2,dalle}.

We provide three sets of samples from our final model with about 2.4 billion parameters.

\paragraph{Completions.} We provide samples for 4 different prompts, without book identity conditioning. There are 16 different samples for each prompt, which were not filtered or cherry-picked. We provide best-effort manual transcriptions in appendix~\ref{apx:sample-transcriptions} (Table~\ref{tab:transcriptions}). These samples demonstrate the ability of the model to maintain intonation and style in most cases. We further note that the majority of the samples are intelligible, and even largely grammatically sound. A smaller number also make sense at a semantic level to some degree: the model often reuses words from the prompt, or words and phrases that are clearly related (e.g. \emph{tree frogs $\rightarrow$ trees, December $\rightarrow$ January, royalty $\rightarrow$ kingdom}).

\paragraph{Unconditional re-ranked samples.} Unconditionally sampling meaningful utterances (without prompts) is a much more challenging task. We drew 512 samples and used the ASR model for intelligibility reranking. We selected the 32 samples with the lowest phoneme prediction entropy. Many of these contain meaningful phrases, though sentences are rarely fully coherent. One sample contains the phrase \emph{`this LibriVox recording is in the public domain'}: this phrase is extremely common in the training data, as it is part of an introduction that precedes every audiobook in the Libri-light dataset.

\paragraph{Book-conditional samples.} We drew samples conditioned on different book embeddings. We selected a few books, drew 16 samples per book, and then manually selected four samples for each. As expected, the topics, vocabulary and style of the utterances are clearly strongly influenced by the conditioning.

\subsection{Variable-rate and fixed-rate representations}
We compare variable-rate discrete speech representations with fixed-rate representations (obtained with a VQ-VAE model) in the context of generative modelling. Transformers trained on fixed-rate representations do not require the various input embeddings that run-length Transformers use, yielding a model that is a bit simpler to implement. On the other hand, their modelling capacity per unit of time is fixed, which may lead to inefficiencies.

We trained a VQ-VAE with a hop size of 256 (relative to the $16 \mathrm{kHz}$ encoder input) and a codebook size of 256. This yields a rate of $62.5 \mathrm{Hz}$, which is slightly lower than the $75 \mathrm{Hz}$ target average event rate we used for the SlowAE reference model. 
However, the effective AER of this model on the Libri-light dataset is actually quite a bit lower (about $51 \mathrm{Hz}$), because this data contains more pauses and variation in length than the SlowAE training data (LibriTTS). We therefore consider this a fair point of comparison.

Both models have similar intelligibility scores (see Figure~\ref{fig:phoneme-accuracies}). In Table~\ref{tab:bps}, we show the average bit rates of different speech representations. We note that the bit rate of the SlowAE event sequences is of the same order of magnitude as that of the VQ-VAE code sequences (which is constant), though higher. This is unsurprising, because ordinal representations are more constrained than categorical ones (see \S\ref{sec:scalar-quantisation}).

We sampled completions from two unconditional models with 1 billion parameters: a vanilla Transformer trained on VQ-VAE code sequences (\emph{fixed-rate}), and an RLT trained on SlowAE event sequences (\emph{variable-rate}). We also used the models to obtain upper bounds for the entropy of the learnt representations, by measuring the negative log-likelihood on the LibriTTS validation set (also in Table~\ref{tab:bps}). The fixed-rate completions have a few issues, such as occasional samples that are completely silent after the prompt, or samples that contain non-speech sounds. We suspect that this is partially due to our use of nucleus sampling with the same threshold ($p = 0.8$) as for RLT models. We experimented with higher thresholds, but found that this negatively affects intelligibility and coherence. That said, these issues clearly are not necessarily inherent to fixed-rate representations.

When discounting these issues, it is clear that both models are able to produce convincing completions in many cases. Note that the variable-rate completions are generally shorter. We sample sequences of length 512 for both settings, and the SlowAE model produces sequences with an above average rate ($\geq 75 \mathrm{Hz}$) for excerpts from the proprietary dataset, so the prompt ends up taking up a larger proportion of the sequence. As a result, it also does a better job of reconstructing the prompts in a phonemically accurate manner.

We conclude that our approach constitutes a useful alternative to VQ-VAE (and other methods that yield fixed-rate discrete sequence representations), which maintains sample quality while enabling adaptive representation sizes. We expect this advantage to be borne out more strongly in more powerful models trained on longer passages (minutes of speech rather than seconds), and we plan to explore this going forward.

\begin{table}
    \centering
    \small
    \begin{tabular}{lc}
    \toprule
    Representation & Average bps \\
    \midrule
    16-bit 24 kHz mono audio & 384,000 \\
    8-bit 24 kHz mono audio (mu-law) & 192,000 \\
    SoTA speech codec~\cite{lyra} & 3,000 \\
    VQ-VAE code sequence & 500 \\
    SlowAE code sequence (LibriTTS, $\sim 75 \mathrm{Hz}$) & 893 \\
    SlowAE code sequence (Libri-light, $\sim 51 \mathrm{Hz}$) & 607 \\
    VQ-VAE entropy bound & $\leq$291 \\
    SlowAE entropy bound (LibriTTS, $\sim 75 \mathrm{Hz}$) & $\leq$435 \\
    SlowAE entropy bound (Libri-light, $\sim 51 \mathrm{Hz}$) & $\leq$296 \\
    \bottomrule
    \end{tabular}
    \caption{Average bit rates (in bits per second) of different speech representations. We also provide upper bounds for the entropy of VQ-VAE code sequences and SlowAE event sequences, which correspond to the negative log-likelihood of 1B parameter Transformer models, measured on the LibriTTS validation set. We provide measurements for two average event rates: the rate of the SlowAE training inputs is approximately $75\mathrm{Hz}$, while the rate of the RLT training inputs is only about $51 \mathrm{Hz}$ for the same model.}
    \label{tab:bps}
\end{table}

\section{Discussion}
\label{sec:discussion}
We discuss a few aspects of our work that merit further consideration, as well as applications and extensions to other domains.

\paragraph{Capabilities.} The unconditional samples produced by RLTs bring to mind the essence of those produced by the popular char-rnn model~\cite{charrnn}: there is a degree of local structure in both the grammatical and semantic sense, but no global coherence. Completion samples also show a rudimentary understanding of the prompt content. This inspires optimism about speech-level language modelling, especially considering where textual language models are now, compared to just half a decade ago.

\paragraph{Evaluation.} While we have attempted to evaluate the representation learning portion of our work both qualitatively and quantitatively, we have not provided a quantitative analysis of RLT sample quality. This is a much more challenging task, which may require a human-in-the-loop approach until convincing computational metrics are developed. \citet{spokenlm} proposed several metrics based on textual transcriptions which are a useful starting point, but these cannot capture aspects of speech which are not shared with written language, so they only provide a partial picture (see \S\ref{sec:language-modelling-in-the-speech-domain}).

\paragraph{Interpretability.} `Events' usually correspond to a change in an underlying state. In SlowAEs, this state is learnt from data, and thus what constitutes an event is defined implicitly. This potentially hampers interpretability. However, if the goal is to use the event sequences as input to other machine learning models, as we have done, this is somewhat inconsequential. That said, it would be interesting to study the structure of the event sequences, beyond our observation that event density correlates with semantically meaningful changes in the nature of the signal (e.g. phoneme identity, see Figure~\ref{fig:scatter-phonemes-events}). This is an important area for further work.

\paragraph{End-to-end learning.} A consistent trend in computer vision, speech processing and natural language processing literature has been to progressively replace more components with learnt models. The underlying idea is that learning from data tends to yield better solutions while requiring less domain knowledge, and every handcrafted component can be a potential barrier limiting the maximal attainable performance.

Our method for learning event-based representations fits this paradigm, but note that we train SlowAEs and RLTs sequentially, rather than jointly in an end-to-end fashion. This is because the objectives of both learning stages are different, and potentially at odds: SlowAEs learn compact representations that allow for faithful speech signal reconstruction, whereas RLTs try to predict how these representations evolve over time as well as possible. If we were to optimise the latter objective for both, clearly the SlowAE representations would collapse, maximising their predictability, but simultaneously minimising their information content. Making integrated end-to-end learning work in this setting is an open question and worth studying further.

\paragraph{Scale and efficiency.} Our speech language models can be improved along two main axes. First, we can scale up the Transformers further (more layers, more trainable parameters), augment them with recent innovations to make them more efficient, and train them on larger, cleaner datasets. While the models we have trained are not small by today's standards, they are still two orders of magnitude smaller than the largest textual language models~\cite{gpt3}, and presumably we would need even larger models to get comparable results in terms of semantic coherence.

Second, we can improve the speech representations, and try to reduce the average event rate required to achieve a particular level of intelligibility. We could do this by incorporating additional inductive biases about speech signals, or perhaps by moving away from reconstruction-based representation learning and using a contrastive approach instead.

While event sequences make abstraction of perceptually unimportant details in speech signals, they still encode timing information exactly, i.e. they are not able to abstract away variations in length. Enabling this could help reduce their information content further, without significantly affecting reconstruction quality.

\paragraph{Applications and other domains.} Besides speech language modelling, variable-rate speech representations could also be useful for other tasks such as text-to-speech, speech recognition, speech-to-speech translation and voice conversion. The approach could also be applied to other modalities which exhibit significant spatiotemporal redundancies, such as music, images and video, though there are some interesting open questions regarding generalisation from sequences to grid- and graph-structured topologies (e.g. what is the analogue of run-length encoding in 2D?). The ideas behind RLTs can be readily adapted for generative models of existing event-based representations, such as those obtained from event cameras~\cite{eventcameras}.

\paragraph{Other considerations.} SlowAE representation values generally increase and decrease quite slowly over time. As a result, we could get fairly high fidelity reconstructions if we encoded the difference in value with the previous event (on the same channel) instead, and restricted the possible values to $\{-1, 0, +1\}$. The resulting sequences look a lot like `spike trains', and reveal a connection with spiking neural networks, which we have not begun to explore.

A strong assumption underlying our work is that the representations we learn should be discrete to be suitable for generative modelling. It would be valuable to be able to relax this requirement, as continuous and differentiable\footnote{Without approximation.} quantities are often easier to work with in neural networks.

Finally, many recent works harnessing discrete representation learning for generative modelling~\cite{challenge,vqvae2,ham,jukebox} use \emph{hierarchically structured} representations as a strong inductive bias. We have not explored this, but it could be one way to achieve abstraction of length variations.

\section*{Acknowledgements}
We would like to thank Johannes Ballé, Mikołaj Bińkowski, Sebastian Borgeaud, Andy Brock, Trevor Cai, Jeffrey De Fauw, Curtis Hawthorne, Jordan Hoffmann, Anna Huang, Geoffrey Irving, Kazuya Kawakami, Pauline Luc, Cyprien de Masson d'Autume, Jacob Menick, Katie Millican, Ben Poole, Alex Pritzel, Jack Rae, Ali Razavi, Roman Ring, Adam Roberts, Evan Shelhamer, Ian Simon and Heiga Zen for ideas, suggestions, discussions and help with implementation. We would like to thank Elizabeth Bamber, Norman Casagrande, Yutian Chen, Ye Jia and Tyler Liechty for their assistance with datasets, as well as Ben Brown, Will Hawkins, Chloe Rosenberg, Eliza Rutherford, Esme Sutherland, Amy Wu and the wider DeepMind team for their support.

We are also thankful to the creators and maintainers of the open source software used in this work, including Python~\cite{python}, NumPy~\cite{numpy}, SciPy~\cite{scipy}, JAX~\cite{jax}, TensorFlow~\cite{tensorflow}, the DeepMind JAX Ecosystem~\cite{dmjax}, librosa~\cite{librosa}, SoX~\cite{sox} and Matplotlib~\cite{matplotlib}.

\bibliography{paper}
\bibliographystyle{icml2021}

\newpage
\appendix
\section{Model architectures and training}
\label{apx:details}

Here, we provide hyperparameters and details about model architectures and training.

\subsection{Slow autoencoders}
\label{apx:slowae-details}

The model we describe is our reference model used to extract representations for RLT training and all further experiments. All other variants we have trained vary in e.g. the type of slowness penalty or quantisation used, or the target AER $R_\mathrm{T}$. Unless otherwise noted, all other aspects of the model are kept unchanged throughout the paper.
 
\begin{figure}[t]
\vskip 0.2in
\begin{center}
\centerline{\includegraphics[width=0.85\columnwidth, trim=0 0 190 310, clip]{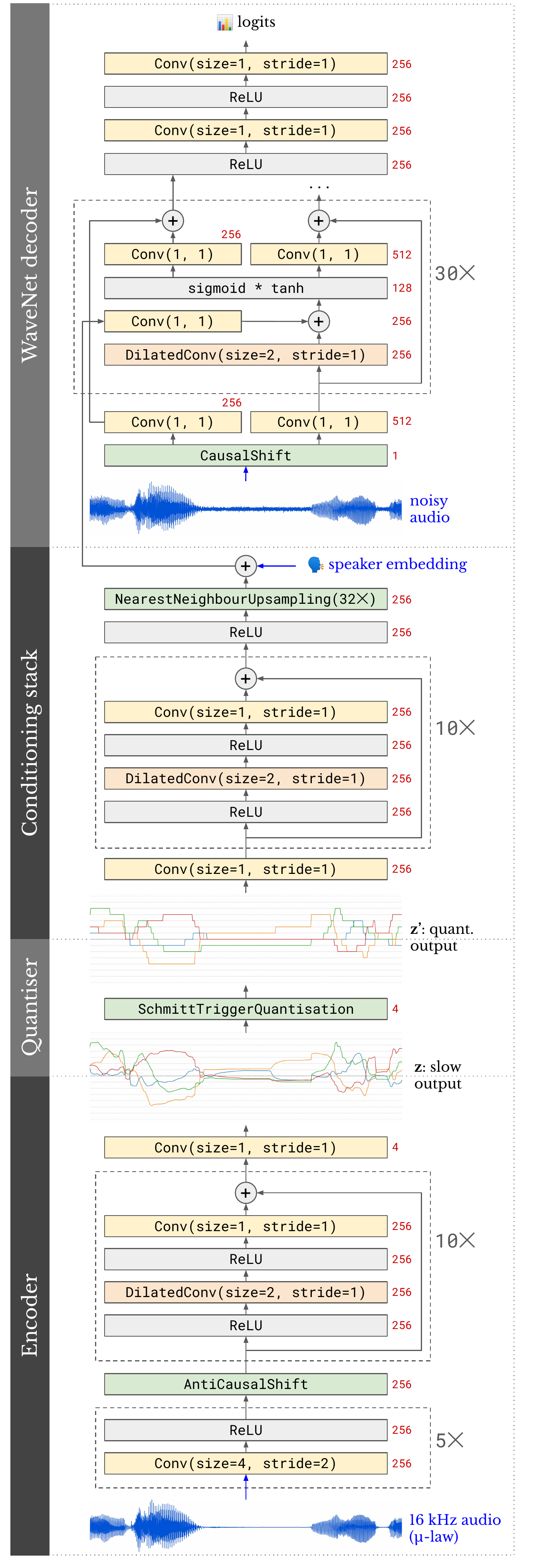}}
\caption{Detailed SlowAE diagram. Dilated convolutions are causal in the WaveNet decoder, and anti-causal in the encoder and conditioning stack. The decoder has 30 residual blocks (3 repetitions of 10 dilation stages), the encoder and conditioning stack have 10 each (2 repetitions of 5 dilation stages). Layer output sizes (number of units) are indicated in {\color{red} red}.}
\label{fig:slowae-diagram}
\end{center}
\vskip -0.2in
\end{figure}

Slow autoencoder models consist of three submodels with trainable parameters (an encoder, a conditioning stack and a decoder) as well as a quantisation module. A detailed diagram is shown in Figure~\ref{fig:slowae-diagram}.

\paragraph{Encoder.} 8-bit mu-law companded audio at 16 kHz is first downsampled in time ($32\times$) using a stack of 5 1D convolutions with filter size 4 and stride 2, alternating with ReLU nonlinearities. The resulting sequence is shifted left by one step, to make the encoder (approximately) anti-causal. A stack of 10 residual blocks with ReLU nonlinearities follows, containing anti-causal dilated convolutions (2 repetitions of 5 dilation stages: $1, 2, 4, 8, 16, 1, 2, 4, 8, 16$). Finally a size-1 convolution produces a continuous output with $C$ channels ($C = 4$ for our reference model). In practice, we implement anti-causality by flipping the encoder input, using causal implementations, and then flipping the output again.

\paragraph{Quantiser.} The encoder output is quantised using Schmitt trigger quantisation (STQ, \S\ref{sec:stq}) with $2k + 1$ levels ($k = 7$ for our reference model).

\paragraph{Conditioning stack.} a single size-1 convolution is followed by a stack of 10 residual blocks with ReLU nonlinearities, with anti-causal dilated convolutions (2 repetitions of 5 dilation stages, same as the encoder). A final ReLU nonlinearity is followed by $32\times$ nearest neighbour upsampling, to obtain a $16 \mathrm{kHz}$ conditioning signal (same resolution as audio).

\paragraph{Decoder.} We use the original conditional WaveNet architecture~\cite{wavenet}, with both residual and skip-connections, and the \texttt{sigmoid * tanh} nonlinearity inside the residual blocks. We use 30 residual blocks with dilated causal convolutions (3 repetitions of 10 dilation stages: $1, 2, \ldots, 1024, 1, 2, \ldots, 1024, 1, 2, \ldots, 1024$).

\paragraph{Training.} We trained all SlowAEs using the Adam optimiser~\cite{adam} ($\beta_1 = 0.9, \beta_2 = 0.999$), with a batch size of 64 and a learning rate of $2\cdot10^{-4}$ for 180,000 steps, $6\cdot10^{-5}$ for 15,000 steps, and finally $2\cdot10^{-5}$ for 5,000 steps (200,000 updates in total). Each training example was a randomly cropped clip of length $5760$ ($360 \mathrm{ms}$) from the LibriTTS dataset~\cite{libritts} (downsampled to $16 \mathrm{kHz}$). We injected Gaussian noise ($\sigma = 0.01$) at the decoder side, before mu-law companding (see \S\ref{sec:slowaes-for-speech}). We kept an exponential moving average of the model parameters during training with decay rate $0.9999$ (\emph{Polyak averaging}, \citet{polyak}).

All models were trained on 8 TPUv3 cores. We set the scale factor of the margin loss $\mu = 100.0$. The scale factor of the slowness penalty $\lambda$ was automatically adapted during training, as described in \S\ref{sec:automatic-penalty-weighting}. For the reference model, the target AER $R_\mathrm{T} = 75 \mathrm{Hz}$.

\subsection{Post-hoc decoders}
\label{apx:post-hoc-details}
Post-hoc decoder training is very similar to training a SlowAE, except that we load the encoder parameters from a checkpoint and keep them fixed (no finetuning). Also, the conditioning stack upsamples to $24 \mathrm{kHz}$ instead of $16 \mathrm{kHz}$, and we do not use speaker embeddings, as there is only one speaker. The decoder gets clean input (no noise injection) and we stop any gradient propagation towards the encoder (using \texttt{jax.lax.stop\_gradient}). We use the proprietary dataset discussed in \S\ref{sec:datasets}, at its original resolution of $24 \mathrm{kHz}$, so training examples of length $5760$ now correspond to $240 \mathrm{ms}$ of audio.

\subsection{Run-length transformers}
\label{apx:rlt-details}

The architecture of our transformer models closely follows that of the decoder from the original paper~\cite{transformer}, with two exceptions: we use learnable relative positional embeddings~\cite{relpos,musictransformer}, and we move the layer normalisation to the input stream of each submodule, so that the residual connections are free of any transformations~\cite{adaptiveinput,languagemultitask,stabilizingrl}.

All models have attention blocks with 8 heads, 2048 units and 512 relative position embeddings. The models with approximately 1 billion parameters have 20 Transformer blocks and a \emph{widening factor} of 4 in the feed-forward blocks (so 8192 hidden units in total), and were trained with a batch size of 256. The models with approximately 2.4 billion parameters have 28 Transformer blocks and a widening factor of 8 in the feed-forward blocks (so 16384 hidden units in total), and were trained with a batch size of 2048.

All models are trained for 1,000,000 steps with the Adam optimiser ($\beta_1 = 0.9, \beta_2 = 0.95$) using a cosine annealing learning rate schedule~\cite{cosineannealing} with warmup: the learning rate starts at $10^{-7}$, increases linearly to $2 \cdot 10^{-4}$ over 1500 steps, and then decreases to $2 \cdot 10^{-5}$ over the course of training. To be able to fit larger models into accelerator memory, we do not use Polyak averaging, and we use optimizer state partitioning as suggested by \citet{zero}. We trained the 1 billion parameter models on 32 TPUv4 cores, and the 2.4 billion parameter models on 128 TPUv4 cores.

\subsection{ASR model for intelligibility evaluation}
\label{apx:asr-details}

We train the ASR model on the proprietary dataset mentioned in \S\ref{sec:datasets}, as it comes with aligned phoneme labels. The provided labels are for 41 different phoneme classes (including silence) at $200 \mathrm{Hz}$. We use the same optimiser settings and learning rate schedule as for SlowAE training (see \S\ref{apx:slowae-details}). We also use Polyak averaging.

The input is $24 \mathrm{kHz}$ audio, which has been mu-law companded and then uncompanded to introduce the typical quantisation noise that is also present in our samples. We train the model on randomly sampled two-second windows ($48000$ timesteps), which are first converted to log-scaled mel-spectrograms with a window length of $240$ and a hop size of $120$, in such a way that the frames are aligned with the phoneme labels at $200 \mathrm{Hz}$. We use 80 mel bins, and the log-scaling is done using $f(x) = \log (1 + 10^5 x) / \log (1 + 10^5)$.

The model itself consists of a size-1 1D convolution (treating the mel bins as input channels) with $512$ units, followed by a stack of residual blocks. Each block contains a ReLU, a size-3 (non-causal) dilated convolution, another ReLU and a size-1 convolution, all with $512$ units. We use 60 blocks (6 repetitions with 10 dilation stages). Finally, a size-1 convolutions produces 41 phoneme class logits for each spectrogram frame. We train this model using the cross-entropy loss, and measure the prediction accuracy averaged across frames on a holdout set.

\section{Implementations}
\label{apx:implementations}

We provide implementations in JAX~\cite{jax} for run-length encoding and decoding, as well as Schmitt trigger quantisation, in Figure~\ref{fig:rle-code}. Note that these implementations do not account for any batch dimensions, but these can be added by using \texttt{jax.vmap}.

\begin{figure*}
\centering
\makebox[\textwidth][c]{%
\begin{minipage}{1.15\textwidth}
\begin{multicols}{2}
\input{rle_code.tex}
\end{multicols}
\end{minipage}}
\caption{JAX implementations of run-length encoding and decoding, as well as Schmitt trigger quantisation.}
\label{fig:rle-code}
\end{figure*}

\section{Negative results}
\label{apx:negative-results}

In this appendix, we informally discuss failed attempts to improve the proposed methods, and alternative approaches that did not pan out. This information is provided to help understand some of our design choices, and to aid researchers and practitioners who are interested in using these methods or investigating ways to improve them. Note that we do not provide thorough ablations, and evidence for these observations is largely anecdotal.

We expressly do not wish to discourage anyone from pursuing the ideas described here. In machine learning, context and details tend to matter a lot, so the fact that these ideas did not work out for us should not be taken to mean that they can never work.

\subsection{Alternative representation learning strategies}
We initially investigated a variable-rate representation learning approach based on VQ-VAE~\cite{vqvae} and byte pair encoding (BPE, \citet{bpe}). This is similar to the strategy used by \citet{discretalk}, who applied SentencePiece encoding~\cite{sentencepiece} to compress VQ-VAE code sequences. However, we found that the compressibility of the code sequences is very limited (relative to the character sequences which BPE and SentencePiece were originally developed for), especially when the encoder and decoder networks are relatively powerful. In practice, the BPE algorithm would learn to group every possible pair of codes into its own `subword unit', before learning any larger units, so the resulting compressed sequences end up being more or less fixed-rate.

This is not too surprising, as the VQ-VAE training objective implicitly tries to maximise the entropy of the code sequences, in order to encode as much information about the input signal as possible. We found the trade-off between compressibility and reconstruction quality quite difficult to control, so eventually we switched our focus to integer quantisation paired with run-length encoding instead.

We also briefly attempted to restrict the `jumps' between subsequent values in SlowAE representations (see Figure~\ref{fig:jump-plots}) to $\{-1, 0, +1\}$ but found that this hampered the initial stages of learning. A possible workaround could be to train a model without this restriction for a while, and then gradually impose it part way through training, but we did not explore this.

\subsection{Reducing event rates}
We discuss a few attempts to further reduce average event rates (AERs) while maintaining the same level of intelligibility, by guiding SlowAEs to encode (or ignore) certain kinds of information in the representations. While we settled on $R_\mathrm{T} = 75 \mathrm{Hz}$, we believe further reductions should be possible.

\paragraph{Augmentations.} In addition to noise injection at the decoder side, we also tried randomly augmenting the decoder audio input, to encourage invariance of the learnt representations to these augmentations. We experimented with pitch shifts, random equalizers, volume changes and temporal jitter, but did not find any of them to be helpful.

\paragraph{Side channels.} We conditioned the decoder on representations of the input signal volume and/or fundamental frequency, so that this information would not be encoded in the slow discrete bottleneck. We also tried adding an additional seperate encoder with a continuous bottleneck which consists of a single temporally-pooled vector, to capture constant signal characteristics. The post-hoc decoder was not conditioned on these signals. None of these meaningfully affected the nature of the learnt discrete representations.

\paragraph{Encoder input.} We downsampled the encoder input audio to lower sample rates ($2, 4, 8 \mathrm{kHz}$), to prevent high frequency details from being encoded. We also tried using mel-spectrogram input instead of raw audio input. All of these negatively affected intelligibility to varying degrees.

\paragraph{Slowness penalty.} We tried various logarithmic variants of the slowness penalty, including some inspired by fat-tailed distributions such as Student's t, in order to encourage burstiness of the learnt representations, but the group-sparse penalty worked much better. We tried additively combining L1 and L2 slowness penalties (the `elastic' slowness penalty, \citet{elasticnet}), but the trade-off was very hard to control, so this ended up behaving mostly like either L1 or L2, depending on the relative weighting. Since SlowAEs are trained entirely by gradient descent, we need not restrict ourselves to convex penalties: we investigated L\nicefrac{1}{2} group sparsity~\cite{lpq}, which looked very promising because it made the encoder outputs much more bursty than regular L1 group sparsity, and allowed much larger jumps. Unfortunately it also negatively affected intelligibility.

\subsection{Alternative RLT architectures}
We experimented with a form of relative attention where the positions were those in the original slow discrete representations, rather than in the event sequence. Because this requires passing in the position information explicitly, we referred to this as \emph{explicit relative attention}. We explored this idea in combination with transformers with multiple `streams' (like XLNet, \citet{xlnet}), but ultimately found that both ideas were not computationally efficient enough on our hardware to be worth it. A vanilla transformer architecture was much easier to scale to billions of parameters, which brings to mind the `bitter lesson' of AI research~\cite{bitterlesson}.

We also tried reweighting the value and length components of the loss: since there are 15 possible values and 256 possible lengths in our setup, and the values are relatively low-entropy, the length component dominates the loss. We wanted to try reducing the importance of getting the lengths exactly right, and potentially introducing more variation in the pacing of the generated speech signals. Contrary to our expectations, this made both the value and the length components worse. Weighting them equally gives the best results in terms of likelihood.

\section{Sample transcriptions}
\label{apx:sample-transcriptions}
We provide best-effort manual transcriptions of the completion samples drawn from our 2.4 billion parameter RLT model. Parts of the samples we deem unintelligible are indicated with `??'. The samples can be listened to at \url{https://vdrl.github.io/}.

\begin{table}
    \scriptsize
    \centering
    \begin{tabular}{ll}
    \toprule
    Prompt & Model output \\
    
    \midrule
    
    \multirow{16}{2cm}{Typical of tree frogs, it has toe pads which enable it to climb smooth vertical...} & sprouts to enable it to \\
    & background. For maybe... \\
    & beautiful tip hat. \\
    & trees. A secondary... \\
    & shallows to ?? \\
    & trees. That is one of... \\
    & tripods and keep ?? \\
    & shelter. But nice... \\
    & trees to back up with... \\
    & trees and ?? \\
    & differences to... \\
    & tree condensers into... \\
    & matter. ?? \\
    & trees answer become... \\
    & bronze ?? \\
    & tree list ?? \\
    
    \midrule

    \multirow{16}{2cm}{WestJet Encore flight 511 on December 24, depart(s)...} & out into the country the first... \\
    & now the testament in January... \\
    & out of Coco's house. \\
    & out of the door. Departing from... \\
    & out of the latter toward the ?? \\
    & again to the height of enemies. \\
    & out of daylight. For Dante's... \\
    & out in a plane stable. \\
    & adequately to their companions of... \\
    & uncongestedly. When... \\
    & out Germany in January. One...\\
    & onto Cloudy Bay. West front. \\
    & off the base contents. With the... \\
    & out to Canada. By ?? \\
    & out on the second day of the week. \\
    & out of the desert in 1870. \\
    
    \midrule

    \multirow{16}{2cm}{Alright, so you have it for later, I've added pencils to your shopping list.} & ?? and third cycle zero. \\
    & I think it has forgiven you entire... \\
    & Wet to make a very large ?? \\
    & And I don't believe Will! So Pennsylvania... \\
    & and ?? the week. But decidedly on... \\
    & There is no place for ?? \\
    & As I did so. And ?? \\
    & and probably only made himself a... \\
    & and sad ?? \\
    & as ?? secretary of Saturday night... \\
    & and you sat beside your finger. I know... \\
    & ?? There must... \\
    & I do not return it to clean ?? \\
    & in a ?? method. What ?? \\
    & ?? that come to me. Perfectly well, but... \\
    & It will do it every time it ?? \\
    
    \midrule

    \multirow{16}{2cm}{Some websites have appropriated lists of royalty, and have reprinted the...} & name privately. The principle shall... \\
    & name of the ?? You have... \\
    & names of the beginning of the present year. But...\\
    & names of the present work of medical science... \\
    & name of contradict each other. And... \\
    & names of the nation. In order to... \\
    & names of ?? all these words ?? \\
    & names of all concerning humanities. They... \\
    & name of ?? \\
    & names of the ?? men. They counter... \\
    & names of the generals of criminals, but... \\
    & names of the government of the kingdom. \\
    & name of \emph{Lapierce} thirteen, ?? \\
    & names of the English represented by... \\
    & names as the bankers' productions. \\
    & names of doctor \emph{Multarelli}'s handwriting. \\
    
    \bottomrule
    \end{tabular}
    \caption{Best-effort manual transcriptions of the completion samples drawn from our 2.4 billion parameter RLT model. Parts of the samples we deem unintelligible are indicated with `??'. We use ellipses to indicate where we would expect the utterance to continue after the end of the sample.}
    \label{tab:transcriptions}
\end{table}

\end{document}